\documentclass[journal]{IEEEtran}
\ifCLASSINFOpdf
\else
\fi
\long\def\comment#1{}

\newfont{\bbb}{msbm10 scaled 700}

\newfont{\bb}{msbm10 scaled 1100}

\newcommand{\RR}{\mbox{\bb R}}

\newcommand{\EE}{\mbox{\bb E}}

\newcommand{\pv}{{\bf p}}

\newcommand{\wv}{{\bf w}}
\newcommand{\vv}{{\bf v}}
\newcommand{\xv}{{\bf x}}

\newcommand{\zv}{{\bf z}}

\newcommand{\Pm}{{\bf P}}

\newcommand{\Tm}{{\bf T}}

\newcommand{\Ac}{{\cal A}}

\newcommand{\Dc}{{\cal D}}

\newcommand{\Nc}{{\cal N}}

\newcommand{\Pc}{{\cal P}}
\newcommand{\Qc}{{\cal Q}}

\newcommand{\Sc}{{\cal S}}

\renewcommand{\arg}{{\hbox{arg}}}

\newcommand{\eqdef}{\stackrel{\Delta}{=}}

\newcommand{\trasp}{{\sf T}}

\usepackage{array, cite}
\usepackage{multirow}
\usepackage{enumerate}
\usepackage{makecell}
\usepackage{tabularx} 
\usepackage{algorithm,algcompatible}
\usepackage{lipsum}
\usepackage{bbm}
\usepackage{times}
\usepackage[cmex10]{amsmath}
\usepackage{algpseudocode}
\usepackage{epsfig,verbatim}
\usepackage{amssymb} 
\usepackage{graphics,subfigure}
\usepackage{xcolor}
\usepackage{graphicx}
\usepackage{tabularx,booktabs,hhline}
\usepackage{color, colortbl}
\definecolor{LightCyan}{rgb}{0.88,1,1}
\definecolor{lightgray}{gray}{0.95}
\usepackage{multirow}
\usepackage{nomencl}
\usepackage{url}
\usepackage{tikz}
\usetikzlibrary{positioning}

\makenomenclature
\nomenclature[1]{Notations}{Descriptions}
\nomenclature[2]{$(\xv_t, y_t)$}{Incoming data samples at time-step $t$}
\nomenclature[3]{$\zv_{i}(\xv)$}{A non-linear feature mapping in (6)}
\nomenclature[3]{$d$}{The dimension of $\xv_t$}
\nomenclature[3]{$s$}{The number of random features}
\nomenclature[4]{$\{{q}_{[\star,i]},{\wv}_{[\star,i]}\}$}{Optimal parameters for RF-OMKL}
\nomenclature[5]{$\{\tilde{q}_{[\star,i]},\tilde{\wv}_{[\star,i]}\}$}{Optimal parameters for expert-based RF-OMKL}
\nomenclature[6]{$\{\tilde{q}_{[t,i]},\tilde{\wv}_{[t,i]}\}$}{Learned parameters via EoKle}
\nomenclature[7]{$\{\check{q}_{[t,i]},\tilde{\wv}_{[t,i]}\}$}{Learned parameters via CoKle}
\nomenclature[8]{$\{\hat{q}_{[t,i]},\hat{\wv}_{[t,i]}\}$}{Learned parameters via BoKle}

\newtheorem{theorem}{Theorem}

\newtheorem{definition}{Definition}

\newtheorem{lemma}{Lemma}

\newtheorem{remark}{Remark}

\newcommand{\argmax}{\operatornamewithlimits{argmax}}
\newcommand{\argmin}{\operatornamewithlimits{argmin}}

\begin{document}
%
\title{On Practical Robust Reinforcement Learning: Adjacent Uncertainty Set and Double-Agent Algorithm}

%
%
%

\author{Ukjo Hwang,~\IEEEmembership{Student,~IEEE,} and~Songnam Hong,~\IEEEmembership{Member,~IEEE}
\thanks{U. Hwang and S. Hong are with the Department of Electronic Engineering, Hanyang University, Seoul, 04763, Korea (e-mail: \{yd1001,snhong\}@hanyang.ac.kr)} 
 }

\maketitle

\begin{abstract}

Robust reinforcement learning (RRL) aims at seeking a robust policy to optimize the worst case performance over an uncertainty set of Markov decision processes (MDPs). This set contains some {\em perturbed} MDPs from a nominal MDP (N-MDP) that generate samples for training, which reflects some potential mismatches between training (i.e., N-MDP) and true environments. In this paper we present an elaborated uncertainty set by excluding some implausible MDPs from the existing sets. Under this uncertainty set, we develop a sample-based RRL algorithm (named ARQ-Learning) for tabular setting and characterize its finite-time error bound. Also, it is proved that ARQ-Learning converges as fast as the standard Q-Learning and robust Q-Learning while ensuring better robustness. We introduce an additional {\em pessimistic} agent which can tackle the major bottleneck for the extension of ARQ-Learning into the cases with larger or continuous state spaces. Incorporating this idea into RL algorithms, we propose {\em double-agent} algorithms for model-free RRL. Via experiments, we demonstrate the effectiveness of the proposed algorithms.
\end{abstract}

\begin{IEEEkeywords}
Reinforcement learning, robust reinforcement learning, robustness, uncertainty set.
\end{IEEEkeywords}

\IEEEpeerreviewmaketitle

\section{Introduction}\label{sec:intro}

In reinforcement learning (RL), learning an agent directly on a real-world system (a.k.a. a testing environment) would be expensive as it requires the large number of data samples. To avoid this complexity, RL algorithms are typically developed from a training environment (a.k.a. a simulator). In the standard RL, it is commonly assumed that the testing environment, on which a learned policy will be deployed, is identical to the training environment that was used to construct the policy \cite{sutton2018reinforcement, szepesvari2010algorithms}. However, in most practical applications, it is often violated due to the simulator modeling errors (i.e., the mismatches between the training and the testing environments), environment changes in real system dynamics over time, and and unpredictable events. As an example, the mass, friction, sensor noise, and floor types in a mobile robot simulator can be different from those in the real world. This mismatch is referred to as the {\em reality gap}, which can degrade the performance of the standard RL algorithms when they are deployed in real-world systems \cite{mannor2007bias, peng2018sim, sunderhauf2018limits}.

To overcome the aforementioned issue, numerous approaches have been developed in the literature. In \cite{iyengar2005robust}, robust adversarial RL (RARL) was proposed, in which the optimization problem to learn a policy is formulated as a zero-sum game. Given this, the so-called adversarial training was developed in  \cite{rajeswaran2016epopt,  pinto2017robust, abdullah2019wasserstein, vinitsky2020robust, hou2020robust}, where an additional adversarial agent modifies a state transition of a training environment so that an original agent ensures the robustness to some potential perturbations.  Although empirical successes have been made, it is still lack of a theoretical understanding. Also, a simple but promising technique to address the reality gap, named domain randomization, was proposed in \cite{tobin2017domain, chen2021understanding}. Instead of training an agent from a single simulator (or training environment), the simulator is randomized during training so that an agent is exposed to a wide range of environments. It is based on the premise that if the variability in the simulator is large enough, the learned agent will generalize to more practical scenarios with no additional training \cite{tobin2017domain}. Last but not least, {\em robust} RL with an uncertainty set (in short, RRL) has been proposed in \cite{nilim2005robust, roy2017reinforcement, badrinath2021robust}, which is closely related to the subject of this paper. The uncertainty set, which plays a key role in guaranteeing the robustness of a learned agent, is defined to be centering at a nominal MDP (N-MDP) (or a training environment) that generate samples for training. Namely, it can include some perturbed MDPs from the N-MDP. We can understand that a testing environment (potentially having some mismatches from the training environment) is uncertain and comes from the uncertainty set. The objective of RRL is to seek a robust policy that optimizes the worst-case performance over the uncertainty set.  Thus, the robust policy can ensure a satisfactory performance when it is deployed in a practical system. Compared to the standard RL, RRL can provide the following advantages:
\begin{itemize}
\item Improved generalization: RRL can be adaptable to various practical scenarios as it is trained to guarantee good performances across diverse environments (or MDPs).
\item Enhanced resilience: Since RRL aims at addressing model uncertainty, perturbations, and unmodeled factors, it can construct a more reliable and resilient agent (called robust agent).
\item Safety and efficiency: RRL not only enhances the safety by reducing the likelihood of undesirable behavior but also improves the resource efficiency by minimizing the required sample size from practical applications. Thus, it is practical and cost-effective.
\end{itemize}

\subsection{Related Works}\label{subsec:related_work}

In this section, we briefly review the existing RRL approaches. The framework of a {\em model-based} RRL (MB-RRL), which is also known as robust MDP (R-MDP), was first introduced in \cite{iyengar2005robust, nilim2005robust}, which can be solved via dynamic programming by exploiting the knowledge of a R-MDP (e.g., a N-MDP and a predefined uncertainty set). This framework has been extensively explored in \cite{xu2010distributionally, wiesemann2013robust, yu2015distributionally, mannor2016robust, petrik2019beyond, lim2019kernel}, developing computationally efficient algorithms. However, these works are limited to the planning problem. MB-RRL algorithms with theoretical guarantees have been proposed in \cite{lim2013reinforcement, tamar2014scaling}, but they are limited to tabular settings or linear function approximations. Although MB-RRL has been investigated using deep neural networks (DNNs), these methods do not provide any theoretical performance guarantee \cite{derman2018soft,derman2020bayesian, mankowitz2019robust, mankowitz2018learning}. These model-based approaches have typical disadvantages as in model-based RLs \cite{berkenkamp2017safe, moerland2023model}. One of the major drawbacks of MB-RRL is to result in model-bias and error, which can degrade the accuracy performance and reliability of a learned agent. Also, MB-RRL is challenging to implement since it requires a large memory to store an estimated model knowledge (e.g., a R-MDP). Moreover, MB-RRL can be limited to the availability of data samples, which can affect the accuracy and validity of the model.

To overcome the aforementioned limitations, a {\em model-free} RRL (MF-RRL) has been recently presented in \cite{wang2021online, wang2022policy, badrinath2021robust, roy2017reinforcement}. In \cite{roy2017reinforcement, badrinath2021robust}, MF-RRL was investigated, where the uncertainty set is developed using confidence region consisting of some constraints on probability distributions. Due to the high-complexity to handle this uncertainty set, MF-RRL algorithm was developed using a proxy uncertainty set with relaxed conditions. However, there is no analytical guarantee for the exactness of such approximation. Very recently in \cite{wang2021online, wang2022policy}, MF-RRL algorithm was developed by introducing the uncertainty set based on Huber's $R$-contamination set \cite{huber1965robust}. Unlike \cite{roy2017reinforcement, badrinath2021robust}, this uncertainty set does not require any condition on probability distributions, which makes it possible to develop MF-RRL algorithm without any relaxation. However, these uncertainty sets have the two major drawbacks (see Section~\ref{subsec:challenges} for details), which makes it hard to be practical.

\subsection{Challenges}\label{subsec:challenges}

We discuss the limitations of the existing MF-RRL frameworks and pose the two key challenges to construct a practical MF-RRL algorithm with a satisfactory performance. In the existing algorithms \cite{wang2021online, wang2022policy, badrinath2021robust, roy2017reinforcement}, the underlying uncertainty sets contain some implausible MDPs (or state transitions), in which a current state can transition to any state in the state space in one step.
Among these transitions, some of them are not plausible in a real world scenario. For example, a drone flying in the air can be pushed by some unexpected air-flow from any direction and with any strength. In this case, the drone can move nearby (i.e., neighboring states) instead of the entire states. This issue is further compounded by the fundamental nature of RRL as it aims to optimize the worst-case performance over the uncertainty set. In other words, specific outliers in the uncertainty set can significantly degrade the performance of MF-RRL algorithm.

In principle, the robust Bellman operator of MF-RRL (as well as MB-RRL) involves a maximization over an uncertainty set, which is induced by the goal of MF-RRL that maximizes the worst-case performance. The computational complexity of solving the maximization depends on the structure of an uncertainty set. Typically, it is intractable to solve the maximization exactly when a N-MDP has large or continuous state spaces. For the existing uncertainty sets \cite{wang2021online, wang2022policy, badrinath2021robust, roy2017reinforcement}, this problem has not been completely addressed. Thus, the existing MF-RRL algorithms are limited to  tabular methods or linear function approximations. More practical MF-RRL algorithms (based on deep neural network (DNN) function approximation) for continuous state spaces have not been well-investigated yet.

The major challenges to develop such a practical MF-RRL are twofold:
\begin{itemize}
\item An uncertainty set should be well-designed to precisely account for practical uncertainties (or perturbations).
\item Given the uncertainty set, a robust policy based on DNN function approximation should be efficiently learned using samples from a training environment in an online fashion. 
\end{itemize}

\subsection{Our Contributions}

To address the above challenges, we first design a new uncertainty set by elaborating Huber's $R$-contamination uncertainty set \cite{wang2021online, wang2022policy} in a more practical way. An additional {\em pessimistic} agent is presented, which makes it possible to construct DNN-based algorithms for MF-RRL with continuous state spaces. We remark that the pessimistic agent has completely different objective than adversarial agents in \cite{rajeswaran2016epopt,  pinto2017robust, abdullah2019wasserstein, vinitsky2020robust, hou2020robust}. Incorporating the idea of the pessimistic agent into the famous DNN-based RL algorithms, we develop {\em double-agent} algorithms consisting of the pessimistic and robust (or original) agents for MF-RRL. Our major contributions are summarized as follows.
\begin{itemize}
    \item We propose an {\em adjacent} $R$-contamination uncertainty set by excluding implausible MDPs (or state transitions) from the $R$-contamination uncertainty set \cite{wang2021online, wang2022policy}. Our uncertainty set contains perturbed MDPs that allow state transitions only to neighboring states (i.e., the states with non-zero transition probabilities from a N-MDP). This set is more reasonable given that the training environment (i.e., the N-MDP) is designed by reflecting a practical environment as close as possible.
    \item We derive the robust Bellman operator for the proposed uncertainty set. Using this, we develop a sample-based RRL algorithm (named ARQ-Learning) for the tabular setting and characterize its finite-time error bound. Also, we prove that it can converge as fast as the standard Q-Learning \cite{li2020sample} and robust Q-Learning \cite{wang2021online} while ensuring better robustness.
    \item A pessimistic agent is additionally introduced, which can solve the key maximization in the robust Bellman operator using samples from the training environment. Combining the idea of the pessimistic agent with the famous RL algorithms (e.g., deep-Q network (DQN) \cite{mnih2015human} and deep deterministic policy gradient (DDPG) \cite{lillicrap2015continuous}), we for the first time develop DNN-based algorithms for MF-RRL with continuous state spaces. Remarkably, our idea can be naturally combined with many other model-free RL algorithms such as soft actor critic \cite{haarnoja2018soft}, twin delayed deep deterministic policy gradient (TD3) \cite{fujimoto2018addressing}, proximal policy optimization (PPO) \cite{schulman2017proximal}, and so on.
    \item Via experiments, it is shown that our algorithms can achieve higher rewards than the standard RL and robust RL algorithms, when there are some perturbations between the training and testing environments. These results can demonstrate the effectiveness of the proposed uncertainty set.
\end{itemize}

\subsection{Outline}

The remaining part of this paper is organized as follows. In Section~\ref{sec:pre}, we formally define RRL framework. In Section~\ref{sec:method}, we introduce a new uncertainty set (named adjacent $R$-contamination uncertainty set) and derive the corresponding robust Bellman operator, which is the key for the design of RRL algorithms. Leveraging the robust Bellman operator, in Section~\ref{sec:ARQ}, we develop a sample-based RRL algorithm (named ARQ-Learning) for the tabular setting and theoretically characterise its finite-time error bound. 
A pessimistic agent is introduced in Section~\ref{sec:FA}, which can tackle the key bottleneck for the extension of ARQ-Learning into larger and continuous state space. As a consequence, PRQ-Learning is proposed for the tabular setting with larger state spaces. 
Combining our idea with the popular RL algorithms such as DQN and DDPG, PR-DQN and PR-DDPG are respectively developed for continuous state spaces. In Section~\ref{sec:exp}, we demonstrate the effectiveness of our algorithms via experiments on RL applications with some perturbations. Section~\ref{sec:con} concludes the paper.

\section{Preliminaries}\label{sec:pre}

We provide some definitions that will be used throughout the paper and formally define a robust reinforcement learning with model uncertainty (in short, RRL).

A Markov decision process (MDP) is defined by a tuple $(\mathcal{S}, \mathcal{A}, \Pm=(p_{s,s'}^a), c, \gamma)$, where $\mathcal{S}$ is the state space, $\mathcal{A}$ is the action space, $\Pm$ is the state transition kernel, $c : \mathcal{S} \times \mathcal{A} \rightarrow \RR$ is the cost function, and $\gamma \in (0, 1)$ is the discount factor.  Here, $p_{s,s'}^a$ represents the probability of transition to state $s' \in \Sc$ when action $a \in \Ac$ is taken at the current state $s \in \Sc$. For ease of notation, we let $p_{s}^a=(p_{s,s'}^{a})$ denote the probability distribution on the possible next state $s' \in \Sc$, given the current state $s$ and action $a$.
At every time step $t$, an agent observes the environment's state $s_t$ and takes an action $a_t$ according to the agent's policy $\pi : \mathcal{S} \times \mathcal{A} \rightarrow [0, 1]$. Then, the environment transitions to the next state $s_{t+1}$ according to $p_{s_t, s_{t+1}}^{a_t}$ and generates a cost $c(s_t, a_t)$. Given a state $s \in \Sc$, the value function of a policy $\pi$ is evaluated as
\begin{equation}\label{eq:value}
V_{\Pm}^{\pi}(s) = \EE_{\pi, \Pm} \left[\sum_{t=0}^{\infty}\gamma^t c(s_t,a_t)| s_0 = s \right],
\end{equation} where $a_t \sim \pi(s_t)$ and $s_{t+1}\sim p_{s_t}^{a_t}$. Given the MDP $(\mathcal{S}, \mathcal{A}, \Pm=(p_{s,s'}^a), c, \gamma)$, the objective of RL is to find an optimal policy  $\pi^{\star}$ such that 
\begin{equation}
\pi^{\star} = \argmin_{\pi} V_{\Pm}^{\pi}(s),\; \forall s \in \Sc.
\end{equation} Also, in model-free RL, the above optimization is solved only using samples from training environment without the knowledge of the underlying MDP.

In robust MDP (R-MDP), there is some uncertainty in an environment for which the state transition kernel $\Pm$ is not fixed but can be changed (or perturbed) at every time step \cite{roy2017reinforcement, badrinath2021robust, wang2021online}. This uncertainty can be captured by an uncertainty set $\mathcal{P}$ containing all possible {\em perturbed} state transition kernels. As in \cite{nilim2005robust, iyengar2005robust, wang2021online}, it is commonly assumed that the uncertainty set has the form of 
\begin{equation}
\Pc = \bigotimes_{(s,a)\in\Sc\times \Ac}\Pc_{s}^{a},
\end{equation} where $\Pc_{s}^{a}$ includes all perturbed transition distributions of a next state $s'$ given the current state $s$ and action $a$. Note that an uncertainty set is defined to be centering at a nominal MDP (N-MDP) $(\Sc,\Ac,\bar{\Pm}=(\bar{p}_{s,s'}^{a}), c, \gamma)$). As aforementioned, in {\em model-free} RRL (MF-RRL), the training environment (defined by an unknown N-MDP) generates samples for training. Let $\Pm_t$ be the state transition kernel at time step $t$ and $\kappa=(\Pm_0,\Pm_1,...)$ be a sequence of state transition kernels. 

Given an uncertainty set $\Pc$, RRL seeks an optimal robust policy $\pi^{\star}$ to minimize the cumulative discounted cost over all state transition kernels in the uncertainty set. In the remaining part of this section, we provide some definitions and notations that will be used to tackle the above problem. Given the state $s$, the {\em robust} state value function of a policy $\pi$ is defined as
\begin{equation}
   V^\pi(s) = \max_\kappa\; \EE_{\pi, \kappa}\left[\sum_{t = 0}^\infty \gamma^t c(s_t, a_t) \middle\vert s_0 = s\right],
\end{equation}
where $\EE_\kappa$ denotes the expectation when state transitions follow $\kappa$. Also, given the state $s$ and action $a$, the {\em robust} action value function of a policy $\pi$ is defined as
\begin{equation}\label{eq:rQ}
    Q^\pi(s, a) = \max_\kappa\; \EE_{\pi, \kappa}\left[\sum_{t = 0}^\infty \gamma^t c(s_t, a_t) \middle\vert s_0 = s, a_0 = a\right].
\end{equation} The optimal policy of the R-MDP is defined as
\begin{equation}
 \pi^{\star} = \argmin_\pi V^\pi(s), \forall s \in \mathcal{S}.
\end{equation} For ease of expositions, we let $V^{\star}$ and $Q^{\star}$ denote the optimal value functions, which implies $V^{\pi^\star}$ and $Q^{\pi^\star}$, respectively. They have the relation such as 
\begin{equation}
V^\star(s) = \min_{a \in \Ac}Q^\star (s, a).
\end{equation} Note that in MF-RRL, the aforementioned optimization is solved only using samples from the training environment.

\begin{remark} One might concern the aforementioned dynamic model, which allows the transition kernel to be time-varying, since the majority of practical settings differs from the training environment in an unpredictable way but fixed throughout the episode. In \cite{iyengar2005robust}, it was shown that the dynamic model is equivalent to the static model, where the transition kernel is static, namely, $\Pm_{t_1}=\Pm_{t_2}$ for any $t_1,t_2$ (i.e., fixed throughout the episode), under a general condition that the agent's policy is stationary and the problem is infinite horizon with a discounted reward. Therefore, an optimal robust policy under the time-varying model is also optimal for the static model.
\end{remark}

%
\section{Proposed RRL Framework}\label{sec:method}

We introduce a new uncertainty set that is more practical and precise than the existing ones. Then, we derive the robust Bellman operator $\Tm$ for the proposed uncertainty set $\Pc$, which will used
as the foundation to develop sample-based algorithms for MF-RRL (see Section~\ref{sec:ARQ} and Section~\ref{sec:FA}).

\subsection{Proposed Uncertainty Set}\label{subsec:US}

In RRL, it is crucial to design an uncertainty set that satisfies the following criteria: i) precisely reflecting practical discrepancies; ii) under this uncertainty set, being manageable to find an optimal robust policy. In the existing works, the two types of uncertainty sets have been widely used:
\begin{itemize}
\item A confidence region-based uncertainty set \cite{roy2017reinforcement, badrinath2021robust}: For $\forall(s, a) \in \Sc \times \Ac$,
\begin{align}
    \mathcal{P}_s^a & \eqdef \left\{\bar{p}_s^a + u \mid u \in \mathcal{U}_s^a \right\},\label{eq:confidence-region}
\end{align} where
\begin{align*}
\mathcal{U}_{s}^a &=\Big\{x \Big| \; \|x\|_2 \leq R,\; \sum_{s \in \mathcal{S}} x_s=0,\\
&\;\;\;\;\;\;\;\;\;\;\; -\bar{p}^{a}_{s, s^\prime} \leq x_{s^{\prime}} \leq 1-\bar{p}^{a}_{s, s^\prime}, \forall s^{\prime} \in \mathcal{S}\Big\}.
\end{align*}
\item $R$-contamination uncertainty set \cite{wang2021online, wang2022policy}: For $\forall(s, a) \in \Sc \times \Ac$,
\begin{equation} \label{eq:r-contamination}
    \mathcal{P}_s^a  \eqdef \left\{(1 - R)\bar{p}_s^a + Rq \mid q \in \Delta_{|\mathcal{S}|}\right\},
\end{equation}where $\Delta_{|\mathcal{S}|}$ is the $(|\mathcal{S}|-1)$-dimensional probability simplex.
\end{itemize} In both sets, the state transition probability $\bar{p}_s^a$ from a N-MDP is adopted as the centroid and a hyperparameter $R\in[0,1]$ denotes a robustness level. This implies that the N-MDP (or the training environment) is designed to be as close as possible to a testing environment and $R$ controls the degree of their mismatches. Based on these uncertainty sets, MF-RRL algorithms were developed in \cite{roy2017reinforcement, badrinath2021robust, wang2021online, wang2022policy}. As identified in Section~\ref{subsec:challenges} and Section~\ref{subsec:comparisons}, however, these uncertainty sets cannot meet the aforementioned two criteria.


This motivates us to present a new uncertainty set as follows:

\begin{definition}\label{def:uncertainty_set}
Given a N-MDP $(\Sc,\Ac,\bar{\Pm}=(\bar{p}_{s,s'}^a),c,\gamma)$ that generates samples for training, we define an {\em adjacent} $R$-contamination uncertainty set: for $\forall(s, a) \in \Sc \times \Ac$,
\begin{equation}\label{eq:proposed_uncertainty}
\mathcal{P}_s^a \eqdef \left\{(1-R)\bar{p}_s^a + R q \mid  q \in \Qc_s \right\},
\end{equation} where $\Qc_s$ contains all {\em feasible} conditional distributions:
\begin{equation}
\Qc_s\eqdef \left\{q \in \Delta_{|\mathcal{S}|} \mid q_{s^{\prime}} = 0, \; \forall s' \in \Sc -  \Nc_s \right\} \subseteq \Delta_{|\Sc|},
\end{equation} and the neighboring set of a state $s\in \Sc$ is defined as 
\begin{equation} \label{eq:neighboring_set}
\Nc_s \eqdef \left\{ s^\prime \in \mathcal{S}\; \middle\vert \; \sum_{a \in \mathcal{A}}\bar{p}_{s, s^\prime}^a \ne 0 \right\}.
\end{equation} Note that when $R=0$, $\Pc$ is the singleton with $\{\bar{\Pm}\}$, namely, in this case, robust RL is reduced to the standard RL.
\end{definition}

\subsection{Robust Bellman Operator} 

When developing RL algorithms, the Bellman operator is the key for policy evaluation steps \cite{sutton2018reinforcement}. Likewise, it is required to define the robust Bellman operator associated with the proposed uncertainty set in \eqref{eq:proposed_uncertainty}. In \cite{nilim2003robustness}, given an uncertainty set $\Pc_s^a$, the robust Bellman operator was derived as 
\begin{equation}
\mathbf{T}Q^{\pi}(s,a) = c(s,a) + \gamma \sigma_{\Pc_{s}^{a}}\left(V^\pi\right),
\end{equation} where $\sigma_{\Pc_{s}^{a}}(\vv) \eqdef \max_{\pv \in \Pc_{s}^{a}}\; \pv^{\trasp}\vv$ is the support function of $\Pc_s^a$. Note that the uncertainty sets in  \eqref{eq:confidence-region}, \eqref{eq:r-contamination}, and \eqref{eq:proposed_uncertainty} can be applied to this Bellman operator. The robust Bellman operator for our uncertainty set in \eqref{eq:proposed_uncertainty} is derived as follows:
\begin{align}
    \mathbf{T} Q^\pi(s, a) 
     &=  c(s,a) + \gamma(1-R)\left[\sum_{s^\prime \in S} \bar{p}_{s, s^\prime}^a V^\pi(s^\prime) \right] \nonumber\\
    &\quad\quad + \gamma R \left[\max _{s'\in \Nc_s }\; V^\pi(s^\prime)\right], \label{eq:bell2}
\end{align} where the neighboring set $\Nc_s$ is defined in \eqref{eq:neighboring_set}.

We theoretically prove the asymptotic convergence of our robust Bellman operator in \eqref{eq:bell2}, namely, the proposed uncertainty set also ensures a theoretical convergence as in the $R$-contamination existing uncertainty set \cite{wang2021online}.

\vspace{0.1cm}
\begin{theorem}
For any $\gamma \in (0, 1)$ and $R \in [0, 1]$, the robust Bellman operator $\mathbf{T}$ in \eqref{eq:bell2} is a contraction with respect to $l_\infty$-norm, and the robust value function $Q^\pi$ has its unique fixed point.
\end{theorem}
\begin{IEEEproof}
The proof is provided in Appendix A.
\end{IEEEproof}

%
\subsection{Comparison}\label{subsec:comparisons}

\begin{figure}
    \centering  
    \includegraphics[width=0.95\linewidth]{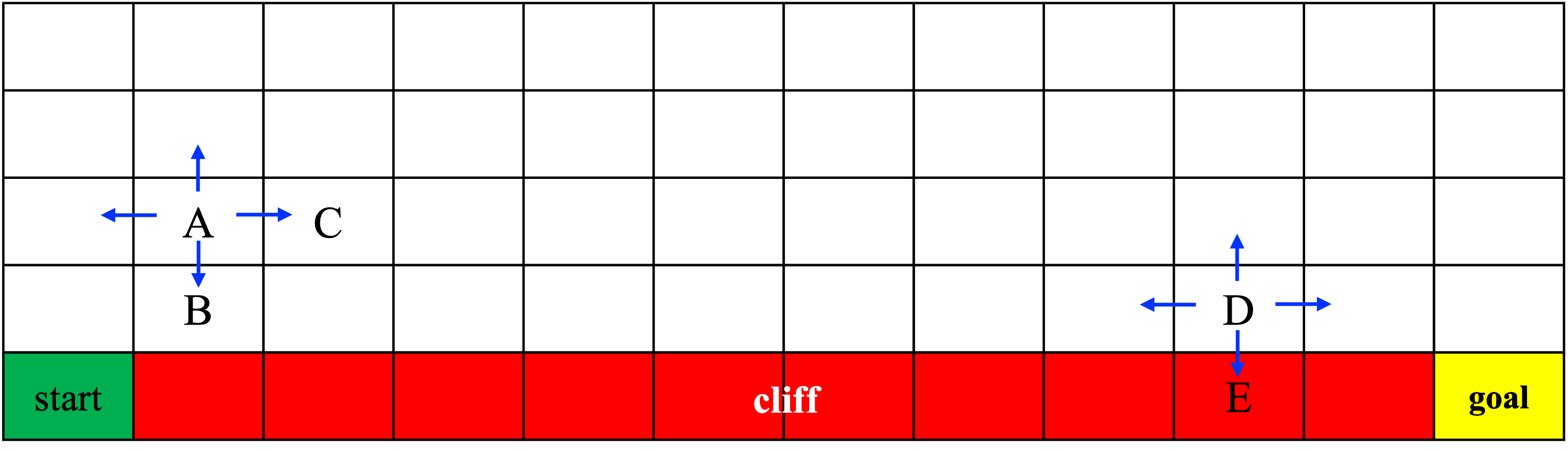} 
    \caption{A grid-world with cliff (or CliffWalking-v0), where agent aims to reach the goal. Here, green represents start state, yellow represents goal, and red represents cliff which gives highest cost. 
    }
    \label{fig:cliff_env}
\end{figure}

We discuss the advantages of our uncertainty set in \eqref{eq:proposed_uncertainty} compared with the state-of-the-art uncertainty set in \eqref{eq:r-contamination}. For completeness, we state the robust Bellman operator based on $R$-contamination uncertainty set in \eqref{eq:r-contamination} as follows:
\begin{align}
    \mathbf{T} Q^\pi(s, a) 
     &=  c(s,a) + \gamma(1-R)\left[\sum_{s^\prime \in \Sc} \bar{p}_{s, s^\prime}^a V^\pi(s^\prime) \right] \nonumber\\
    &\quad\quad + \gamma R \left[\max _{s'\in \Sc }\; V^\pi(s^\prime)\right]. \label{eq:r-contamination-bell}
\end{align} In comparison with \eqref{eq:bell2}, it is noticeable that there is no state-dependent constraint for the maximization in \eqref{eq:r-contamination-bell}. Specifically, the last terms in \eqref{eq:bell2} and \eqref{eq:r-contamination-bell} play a crucial role in guaranteeing the robustness by updating the robust action value function in a more conservative way. Noticeably, these terms are quite different: the former is a state-dependent while the latter is constant regardless of a current state $s$. To clearly explain the impact of such difference, we provide an example with Fig.~\ref{fig:cliff_env}. Obviously, the state `E' tends to have a higher $V^{\pi}(s=\mbox{`E'})$ as it is in the cliff (i.e., the dangerous state). For the update of $Q^{\pi}(\mbox{`D'},a)$, it is reasonable to reflect the danger of the neighboring state `E' since with high probability, the state `D' can transition to the neighboring state `E' by some unexpected events in the testing environment. On the other hand, it is not the case when updating  $Q^{\pi}(\mbox{`A'},a)$ as the state `A' is far from the state `E' (i.e., this transition is not possible in one step). Nevertheless, in \eqref{eq:r-contamination-bell}, such implausible transition is taken into account. Moreover, when $V^{\pi}(\mbox{`E'})$ becomes extremely high, which is highly likely to occur, the maximum state $\mbox{`E'}=\argmax_{s^\prime \in \Sc}V^\pi(s^\prime)$ is unchanged and accordingly, an almost same value is added at every time step. This issue is due to the state-independent maximization in \eqref{eq:r-contamination-bell}. As shown in \eqref{eq:bell2}, the proposed uncertainty set can address the aforementioned drawbacks by restricting the search space of the maximization to a neighboring set. It makes sense to use the neighboring set based on the premise that transitions that are considered infeasible in N-MDP are more likely to remain infeasible in a test environment. Therefore, our uncertainty set is more elaborated in a practical way than the existing one in \eqref{eq:r-contamination}. 

\vspace{0.2cm}
\begin{remark} 
One might concern that for a certain environment, a state $s$ can transition to any state $s' \in \mathcal{S}$ in one step. Thus, the existing uncertainty sets in \eqref{eq:confidence-region} and \eqref{eq:r-contamination} are more suitable. However, for this environment, the corresponding neighboring set in \eqref{eq:neighboring_set} becomes the entire state space (i.e.,  $\mathcal{N}_s = \mathcal{S}$). Our uncertainty set in \eqref{eq:proposed_uncertainty} is naturally reduced to the $R$-contamination uncertainty set in \eqref{eq:r-contamination}. We remark that the proposed uncertainty set is not a special case of the existing one in \eqref{eq:r-contamination}  but a generalized version in a more practical way.
\end{remark}
\vspace{0.2cm}

\section{Proposed ARQ-Learning: Tabular Case}\label{sec:ARQ}
Focusing on the tabular case with finite state and action spaces, we develop a sample-based RRL algorithm based on the robust Bellman operator in \eqref{eq:bell2}. The proposed method is referred to as {\bf A}djacent {\bf R}obust {\bf Q}-Learning (ARQ-Learning). Strictly speaking, ARQ-Learning can be considered as a model-based algorithm as it needs to estimate the partial information of a model such as the neighboring sets $\hat{\Nc}_s$. Genuine MF-RRL algorithms will be developed in Section~\ref{sec:FA}.

We first estimate the neighboring set $\Nc_s, \forall s \in \Sc$ using samples from the training environment. During training, if a current state $s$ transitions to a state $s'$, then $s'$ is included to the neighboring set $\hat{\Nc}_{s}$. This process is called environment setup in Algorithm 1. By construction, we have $\hat{\Nc}_s \subseteq \Nc_s$. After sufficient time steps, it is expected that $\hat{\Nc}_s$ converges to the neighboring set $\Nc_s$. Based on our theoretical analysis in Theorem 2, it is sufficient that $\Nc_s \subseteq \hat{\Nc}_s$ contains the state to maximize the optimal value function, i.e., 
\begin{equation}
    \argmax_{s' \in \Nc_s}V^{\star}(s') \in \hat{\Nc}_s,
\end{equation} which can hold with high probability due to the nature of RL training. Leveraging the estimated neighboring sets and the robust Bellman operator in \eqref{eq:bell2}, the update of ARQ-Learning with the sample $(s, a, c, s^\prime) \sim \mathcal{D}$ is derived as follows:
\begin{align}
    &Q^\pi(s,a) \leftarrow  (1-\alpha)Q^\pi(s,a) \label{eq:ARQ-update}\\ 
    &+\alpha\left(c +  \gamma (1-R)V^\pi(s^\prime) + \gamma R \max_{s^\prime \in \hat{\Nc}_{s}}V^\pi(s^\prime) \right),\nonumber
\end{align} where $\alpha>0$ is a learning rate. The overall procedures of ARQ-Learning are provided in {\bf Algorithm~\ref{alg:arq}}.

\begin{algorithm}[t]
    \caption{ARQ-Learning} \label{alg:arq}
    \begin{algorithmic}[1] 
        \State {\bf Input:} Learning rate $\alpha>0$, discounted factor $\gamma\in(0,1)$, $\hat{\Nc}_s = \phi, \forall s \in \Sc$, and robustness-level $R$.
        \State {\bf Initialization:} $Q^\pi(\cdot,\cdot)$, $s_0 \in \mathcal{S}$.
            \For{each epoch}
                \For{each environment step}
                    \State $a_t \sim \pi(s_t)$ and $s_{t+1} \sim \bar{p}_{s_t}^{a_t}$
                    
                    \State $\mathcal{D} \leftarrow \mathcal{D} \cup \{ (s_t ,a_t, c_t, s_{t+1}) \}$
                    \State $\hat{\Nc}_{s_t} \leftarrow \hat{\Nc}_{s_t} \cup \{s_{t+1}\}$ and $s_t \leftarrow s_{t+1}$
                \EndFor
                \For{each update step}
                    \State {\tiny$\bullet$} Compute the state value
                    
                    $\qquad (s, a, c, s^\prime) \sim \mathcal{D}$
                    
                    $\qquad V_t^\pi(s') = \min_{a' \in \Ac} Q_t^\pi(s', a')$
                    
                    \State {\tiny$\bullet$} Update $Q^{\pi}(s,a)$ via \eqref{eq:ARQ-update}
                \EndFor
            \EndFor
    \end{algorithmic}
\end{algorithm}

We conduct a theoretical analysis to derive the finite-time error bound of the proposed ARQ-Learning. Our analysis is some extension of the prior work in \cite{wang2021online} to fit to the proposed uncertainty set. Before stating the main result, we define the useful notations and assumptions. Let $\mu_{\pi_{b}}$ be the stationary distribution over $\mathcal{S} \times \mathcal{A}$ induced by behavior policy $\pi_{b}$. Given $\mu_{\pi_{b}}$, we define 
\begin{align}
    \mu_{\min }&=\min _{(s, a) \in \mathcal{S} \times \mathcal{A}} \mu_{\pi_{b}}(s, a)\\
    t_{\rm mix}&=\min \left\{t: \max _{s \in \Sc, a \in \Ac} d_{\mathrm{TV}}(\mu_{\pi_{b}}, p_{s}^{a} ) \leq 0.25\right\}, 
\end{align} where $d_{\mathrm{TV}}$ denotes the total variation distance. 

\vspace{0.1cm}
\noindent{\bf Assumption 1.} The Markov chain induced by the stationary behavior policy $\pi_b$ and the state transition kernel $\bar{p}_s^a, \forall (s,a) \in \mathcal{S}\times \mathcal{A}$ is uniformly ergodic.

\noindent This assumption is commonly used in the analysis of the standard Q-Learning \cite{li2020sample} and the existing robust Q-Learning in \cite{wang2021online}. 

\vspace{0.1cm}
\noindent{\bf Assumption 2.} For any state $s \in \Sc$, an estimated neighboring set $\hat{\Nc}_s \subseteq \Nc_s$ contains a state to maximize the optimal value function, i.e., 
\begin{equation*}
    \argmax_{s^\prime \in \mathcal{N}_s} V^{\star}(s^\prime) \in \hat{\mathcal{N}}_s.
\end{equation*}

\begin{theorem}\label{thm:error_bound} Under Assumption 1 and Assumption 2, there exist some constants $c_0, c_1 > 0$ such that for any $0 < \delta<1$ and $0 < \varepsilon \le \frac{1}{1-\gamma}$, the proposed ARQ-Learning in Algorithm 1 with learning rate $\alpha =\frac{c_1}{\log \left(\frac{|S||\mathcal{A}|T}{\delta}\right)} \min \left(\frac{1}{t_{\text {mix }}}, \frac{(1-\gamma)^4\varepsilon^2}{\gamma^2}\right)$ satisfies the following bound with probability at least $1-\delta$:
\begin{equation*}
    \left|Q_T(s, a)-Q^{\star}(s, a)\right| \leq \varepsilon,\; \forall (s, a) \in \mathcal{S} \times \mathcal{A},
\end{equation*} provided that $T$ satisfies 
\begin{align*}
        T &\geq \frac{c_0}{\mu_{\min}}\left(\frac{1}{(1-\gamma)^5 \varepsilon^2}+ \frac{t_{\text {mix }}}{(1-\gamma)}\right) \\
        &\quad\quad\quad\quad\quad \times \log \left(\frac{|\mathcal{S}||\mathcal{A}|T}{\delta}\right) \log \left(\frac{1}{(1-\gamma)^2\varepsilon}\right).
    \end{align*}
\end{theorem}
\begin{IEEEproof}
    The proof is provided in Appendix B.
\end{IEEEproof}

\vspace{0.1cm}
From Theorem~\ref{thm:error_bound}, we can identify that a sample size $\tilde{\mathcal{O}}\left(\frac{1}{\mu_{\min }(1-\gamma)^5 \varepsilon^2}+ \frac{t_{\operatorname{mix}}}{\mu_{\min }(1-\gamma)}\right)$ is required to guarantee an $\varepsilon$-accuracy. The complexity of the proposed ARQ-Learning is well-matched with those of the standard Q-learning in \cite{li2020sample} and robust Q-Learning in \cite{wang2021online} while ensuring the robustness to practical model uncertainties. Despite its effectiveness, ARQ-Learning can suffer from the large memory space to store the estimated neighboring set $\hat{\Nc}_{s}, \forall s\in \Sc$ especially when the size of $\Sc$ becomes larger. In fact, this is the major bottleneck to develop RRL algorithm for large or continuous state spaces. This problem will be addressed in the subsequent section.


\begin{algorithm}[htb!]
\caption{Sampling}\label{alg:p-sampling}
\begin{algorithmic}[1]
\For{each environment step}
    \State {\tiny$\bullet$} Perform the state sharing as $x_t \leftarrow s_t$
    
    \State {\tiny$\bullet$} Sampling by the robust agent
    
    $\qquad a_t \sim \pi(s_t)$, $s_{t+1} \sim \bar{p}_{s_t}^{a_t}$, and $c_t = c(s_t,a_t)$

    \State {\tiny$\bullet$} Sampling by the pessimistic agent
    
    $\qquad u_t \sim \phi(x_t) $, $x_{t+1} \sim \bar{p}_{x_t}^{u_t}$, and $c_t^p = - c(x_t,u_t)$

    \State {\tiny$\bullet$} Store samples in the replay buffer
    
    $\qquad \mathcal{D} \leftarrow \mathcal{D} \cup \{ (s_t ,a_t, c_t, s_{t+1}, u_t,  c_t^p , x_{t+1}) \}$

    \State {\tiny$\bullet$} Update the state as $s_t \leftarrow s_{t+1}$
    
\EndFor
\end{algorithmic}
\end{algorithm}

\begin{algorithm}[t]
\caption{PRQ-Learning}\label{alg:prq}
\begin{algorithmic}[1]
\State {\bf Input:} Learning rate $\alpha>0$, discounted factor $\gamma\in(0,1)$, and robustness-level $R$.
\State {\bf Initialization:} $Q^\pi(\cdot,\cdot)$, $Q^\phi(\cdot,\cdot)$, $s_0 \in \mathcal{S}$.
    \For{each epoch}
            
            
        \State {\tiny$\bullet$} Sampling via Algorithm \ref{alg:p-sampling}.
        \For{each update step}
            \State {\tiny$\bullet$} Compute the state values
            
            $\qquad (s, a, c, s^\prime, u, c^p, x^\prime) \sim \mathcal{D}$
            
            $\qquad V_t^\pi(x^\prime) = \min_{a' \in \Ac} Q_t^\pi(x^\prime, a')$
            
            $\qquad V_t^\pi(s^\prime) = \min_{a' \in \Ac} Q_t^\pi(s^\prime, a')$
            
            $\qquad V_t^\phi(x^\prime) = \min_{u' \in \Ac} Q_t^\phi(x^\prime, u')$
            
            \State {\tiny$\bullet$} Update $Q^{\phi}(s,u)$ via \eqref{eq:PRQ-pessimistic} and $Q^{\pi}(s,a)$ via \eqref{eq:PRQ-robust}
        \EndFor
    \EndFor
\end{algorithmic}
\end{algorithm}

%
\section{Proposed MF-RRL Algorithms: \\DNN Function Approximations}\label{sec:FA}

In this section, we extend the proposed ARQ-Learning into more practical cases that state and action spaces are larger or even continuous. This extension is enabled by means of a pessimistic agent. Combining the idea of the pessimistic agent with the popular RL methods, we develop {\em double-agent} algorithms for MF-RRL with large or continuous state spaces. 

\begin{figure}
    \centering
    \begin{tikzpicture}[
    SIR/.style={circle, draw = black}
    ]
    
    \node[SIR] (s0) {$s_0$};
    \node[SIR] (s1) [right=of s0, xshift=0.5cm] {$s_1$};
    \node[SIR] (s2) [right=of s1, xshift=0.5cm] {$s_2$};
    \node[SIR] (s3) [right=of s2, xshift=0.5cm] {$s_3$};
    
    \node[SIR] (x1) [below=of s1] {$x_1$};
    \node[SIR] (x2) [below=of s2] {$x_2$};
    \node[SIR] (x3) [below=of s3] {$x_3$};

    \draw[->] (s0.east) to node[above] {$\pi(s_0)$} (s1.west) ;
    \draw[->] (s1.east) to node[above] {$\pi(s_1)$} (s2.west);
    \draw[->] (s2.east) to node[above] {$\pi(s_2)$} (s3.west);
    
    \draw[->] (s0.east) .. controls +(down:7mm) and +(left:7mm) .. node[above, xshift=0.3cm] {$\phi(s_0)$} (x1.west);
    \draw[->] (s1.east) .. controls +(down:7mm) and +(left:7mm) .. node[above, xshift=0.3cm] {$\phi(s_1)$} (x2.west) ;
    \draw[->] (s2.east) .. controls +(down:7mm) and +(left:7mm) .. node[above, xshift=0.3cm] {$\phi(s_2)$} (x3.west);
        
    \end{tikzpicture}
    \caption{Diagram of the states of the robust and pessimistic agents ($s$ and $x$, respectively). $\pi$ and $\phi$ represent the policies of the robust and pessimistic agents, respectively.}
    \label{fig:state_diagram_sum}
\end{figure}
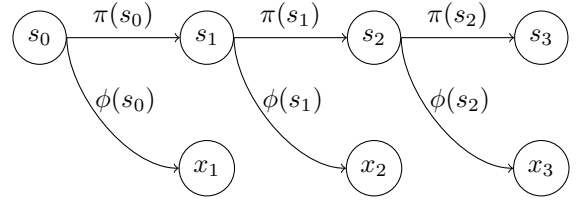

%
\subsection{The Proposed Pessimistic Agent}\label{subsec:pessimistic-agent}

In the robust Bellman operator \eqref{eq:bell2}, the maximization over neighboring states should be addressed. Although the neighboring set $\Nc_s$ is given a priori, it is challenging to solve the maximization when the state space is large or continuous. To overcome this, we propose an additional pessimistic agent whose goal is to find the solution of the maximization in \eqref{eq:bell2}, i.e., it can be thought of as max-solver.

From here on, to distinguish from the state $s \in \Sc$ and action $a\in \Ac$ of a robust agent, we let  $x \in \Sc$ and $u \in \Ac$ be the state and action of a pessimistic agent. Since the goal of the pessimistic agent is to efficiently solve the maximization in \eqref{eq:bell2} which needs to find the maximum cost value among neighboring states, the pessimistic agent is trained to find a policy that maximizes the cumulative discounted cost. Thus, the cost function of the pessimistic agent is determined as the negative of the original cost function $c(x,u)$, i.e., $c^p(x,u)= - c(x,u)$. We also let $\phi$ denote the policy of the pessimistic agent. To train the pessimistic agent, the value functions of the policy $\phi$ are defined as
\begin{align*}
   V^\phi(x) &= \EE_{\phi, \Pm}\left[\sum_{t = 0}^\infty \gamma^{t} c^p(x_t, u_t) \middle\vert x_0 = x\right]\\
   Q^\phi(x, u) &= \EE_{\phi, \Pm}\left[\sum_{t = 0}^\infty \gamma^{t} c^p(x_t, u_t) \middle\vert x_0 = x, u_0 = a\right].
\end{align*} The optimal policy of the pessimistic agent is obtained as
\begin{equation}
   \phi^{\star} = \argmin_\phi V^\phi(x),\; \forall x \in \mathcal{S}.
\end{equation} 

As explained in Remark~\ref{rem:state_sharing} below, we employ the {\em state sharing}, where at the beginning of sampling, the current state of the pessimistic agent is replaced with that of the robust agent such as
\begin{equation}
   x \leftarrow s.
\end{equation}
Subsequently, the pessimistic agent takes an action $u \in \Ac$ and accordingly, the sample $(x = s, u, c^p(x, u), x^\prime)$ is generated. Similarly, the robust agent generates the sample $(s, a, c(s, a), s^\prime)$. Ultimately, all these samples are collectively stored in the replay buffer $\mathcal{D}$:
\begin{equation}
   \mathcal{D} \leftarrow \mathcal{D} \cup \{ (s = x ,a, c(s, a), s^\prime, u, c^p(x, u), x^\prime) \}.
\end{equation} Fig.~\ref{fig:state_diagram_sum} provides insight into how the ultimate trajectories of the robust and pessimistic agents are structured. Also, the detailed descriptions of the sampling by the robust and the pessimistic agents are given in Algorithm~\ref{alg:p-sampling}. Using these samples, the robust Bellman equation in \eqref{eq:bell2} can be reformulated as follows:
\begin{align}
   \mathbf{T} Q^\pi(s, a) 
    &=  c(s,a) + \gamma(1-R)\left[\sum_{s^\prime \in S} \bar{p}_{s, s^\prime}^a V^\pi(s^\prime) \right] \nonumber\\
   &\quad\quad + \gamma R \left[V^\pi(x^\prime)\right]. \label{eq:bell3}
\end{align}
We highlight that although the above robust Bellman equation is used to train the robust agent, it is crucial to acknowledge that the sample $x'\in\Sc$ provided by the pessimistic agent is also incorporated into the equation. The reformulation in \eqref{eq:bell3} is possible primarily because sampling with the pessimistic agent is based on the training environment (or N-MDP). This ensures that the sample $x^\prime$ generated by the pessimistic agent is guaranteed to exist within the neighboring set such as
\begin{equation}  \label{eq:pess_condition1}
    x^\prime \in \Nc_s.
\end{equation} Furthermore, since the pessimistic agent is trained for the purpose of maximizing the cost, the following equation holds with high probability:
\begin{equation} \label{eq:pess_condition2}
    V^\pi(x^\prime) \approx \max _{s'\in \Nc_{s} } V^\pi(s^\prime).
\end{equation} Remarkably, the robust Bellman equation in \eqref{eq:bell3} no longer involves the maximization problem and instead, it can be solved using samples exclusively from the replay buffer. Therefore, based on this, we can develop various sample-based algorithms for MF-RRL with larger or continuous state spaces. The resulting algorithms are provided in the subsequent subsection.

\vspace{0.2cm}
\begin{remark}\label{rem:state_sharing}
A pessimistic agent aims at maximizing the cumulative cost as the opposite of robust agent. As manifested in RL, the maximum cost tends to terminate an episode early, thereby hindering the pessimistic agent from exploring states sufficiently. This makes pessimistic agent to be poorly trained. We address this problem by means of the {\em state-sharing} technique, wherein the current state of pessimistic agent is coupled with that of robust agent. Harnessing the exploration ability of robust agent, pessimistic agent can be trained in the following ways: at every time step, before interacting with the training environment, the current state of pessimistic agent is updated as that of robust agent, i.e., $x_t = s_t$. Fig. 2 provides the insight into how the ultimate trajectories of the robust and pessimistic agents are structured. Because of this state-sharing, in \eqref{eq:PRQ-pessimistic}, the current state $x_t$ is replaced with $s_t$. The detailed procedures are provided in Algorithm~\ref{alg:p-sampling}.
\end{remark}
\vspace{0.2cm}

\subsection{Double-Agent Algorithms}

We develop double-agent algorithms by incorporating the idea of a pessimistic agent in Section~\ref{subsec:pessimistic-agent} into the well-established RL algorithms such as Q-Learning \cite{sutton2018reinforcement}, deep-Q network (DQN) \cite{mnih2015human}, and deep deterministic policy gradient (DDPG) \cite{lillicrap2015continuous}. The resulting methods are named {\bf P}essimistic {\bf R}obust {\bf Q}-Learning (PRQ-Learning), {\bf P}essimistic {\bf R}obust DQN (PR-DQN), and {\bf P}essimistic {\bf R}obust DDPG (PR-DDPG), respectively. For these methods, the sampling method by the robust and the pessimistic agents is outlined in Algorithm 2. Thus, it is noteworthy that the following samples are obtained from the replay buffer:
\begin{equation}
    (s=x, a, c, s', u, c^p, x') \sim \mathcal{D}.
\end{equation} The overall procedures of PRQ-Learning, PR-DQN and PR-DDPG are described in  Algorithm \ref{alg:prq}, Algorithm \ref{alg:pr_dqn}, and Algorithm \ref{alg:pr_ddpg}, respectively.

\subsubsection{PRQ-Learning}\label{subsec:PRQ}

As explained in Section~\ref{subsec:pessimistic-agent}, training the pessimistic agent follows the same procedure as the standard RL with a modified cost function.  Thus, the Q-Learning update for the pessimistic agent can be derived as follows:
\begin{equation}\label{eq:PRQ-pessimistic}
    Q^{\phi}(x,u) \leftarrow (1-\alpha)Q^{\phi}(x, u)  + \alpha \left(c^p+V^{\phi}(x')\right),
\end{equation}
where $\alpha>0$ is a learning rate. 
Also, using the robust Bellman operator in \eqref{eq:bell3}, the Q-Learning update for the robust agent is obtained such as
\begin{align}
    &Q^\pi(s,a) \leftarrow  (1-\alpha)Q^\pi(s,a) \nonumber \\ 
    & \quad+\alpha(c+ \gamma R \cdot V^\pi(x') + \gamma (1-R)V^\pi(s')).\label{eq:PRQ-robust}
\end{align} Note that the sample $x'$ is acquired from the pessimistic agent. Leveraging the updates in \eqref{eq:PRQ-pessimistic} and \eqref{eq:PRQ-robust}, PRQ-Learning is developed (see Algorithm~\ref{alg:prq} for the detailed procedures). We remark that PRQ-Learning is an indeed model-free algorithm, whereas the existing Q-Learning based methods in \cite{roy2017reinforcement, badrinath2021robust, wang2021online, wang2022policy} as well as ARQ-Learning require additional and demanding computations to solve the maximization in the robust Bellman operators.

\subsubsection{PR-DQN} We construct the robust Q-network and the pessimistic Q-network for the robust and the pessimistic agents, respectively. The loss function of the pessimistic Q-network, denoted by $Q^{\phi}(x,u;\psi)$, defined as follows:
\begin{align}\label{eq:pr-dqn-p}
    L_{Q^\phi}(\psi)= \EE_{(x, u, c^p, x^\prime) \sim \Dc}\left[(Q^\phi(x, u ; \psi) - y)^2\right],
\end{align} where $y = c^p + \gamma \bar{V}^\phi(x^\prime)$ and $\bar{V}^\phi$ is a target network. Also, the loss function of the robust Q-network, denoted by $Q^{\pi}(s,a;\omega)$, is defined as follows:
\begin{align}\label{eq:pr-dqn-r}
        L_{Q^\pi}(\omega) &= \EE_{(s, a, c, s^\prime, x^\prime) \sim \Dc}\left[(Q^\pi(s, a;\omega) - y)^2 \right],
\end{align} where $y= c + \gamma(1 - R)\bar{V}^\pi(s^\prime)  + \gamma R \cdot \bar{V}^\pi(x^\prime) $ and $\bar{V}^\pi$ is a target network. The updates for the target networks $\bar{V}^\phi$ and $\bar{V}^\pi$ follows the same procedures as the standard DQN \cite{mnih2015human}, which are given in Algorithm~\ref{alg:pr_dqn}. Likewise PRQ-Learning, both Q-networks are trained only using samples from the training environment.

\begin{algorithm}[htb!]
\caption{PR-DQN}\label{alg:pr_dqn}
\begin{algorithmic}[1]
\State {\bf Input:} Learning rate $\lambda_Q>0$, discounted factor $\gamma\in(0,1)$, robustness-level $R$, and $\tau > 0$.
\State {\bf Initialization:} $\psi$, $\bar{\psi}$, $\omega$, $\bar{\omega}$, $s_0 \in \mathcal{S}$.
    \For{each epoch}
            
            
            

            

            

            
        \State {\tiny$\bullet$} Sampling via Algorithm~\ref{alg:p-sampling}
        \For{each update step}
            \State {\tiny$\bullet$} Sampling from the replay buffer
            
            $\qquad (s, a, c, s^\prime, u, c^p, x^\prime) \sim \mathcal{D}$
            
            \State {\tiny$\bullet$} Compute the state values 
            
            $\qquad\bar{V}^\pi(x^\prime) = \min_{a^\prime \in \Ac} \bar{Q}^\pi(x^\prime, a^\prime)$
            
            $\qquad\bar{V}^\pi(s^\prime) = \min_{a^\prime \in \Ac} \bar{Q}^\pi(s^\prime, a^\prime)$
            
            $\qquad\bar{V}^\phi(x^\prime) = \min_{u^\prime \in \Ac} \bar{Q}^\phi(x^\prime, u^\prime)$

            \State {\tiny$\bullet$} Update the Q-networks
            
            $\qquad\psi \leftarrow \psi-\lambda_{Q} \nabla_{\psi} L_{Q^\phi} \left(\psi\right)$
            
            $\qquad\omega \leftarrow \omega-\lambda_{Q} \nabla_{\omega} L_{Q^\pi} \left(\omega\right)$

            \State {\tiny$\bullet$} Update the target networks
            
            $\qquad\bar{\psi} \leftarrow \tau \psi+(1-\tau) \bar{\psi}$
            
            $\qquad\bar{\omega} \leftarrow \tau \omega+(1-\tau) \bar{\omega}$
        \EndFor
    \EndFor
\end{algorithmic}
\end{algorithm}

\begin{algorithm}[htb!]
\caption{PR-DDPG}\label{alg:pr_ddpg}
\begin{algorithmic}[1]
\State {\bf Input:} Learning rate $\lambda_Q>0$, $\lambda_p > 0$, discounted factor $\gamma\in(0,1)$, robustness-level $R$, and $\tau > 0$.
\State {\bf Initialization:} $\theta$, $\bar{\theta}$, $\sigma$, $\bar{\sigma}$, $\psi$, $\bar{\psi}$, $\omega$, $\bar{\omega}$, $s_0 \in \mathcal{S}$.
    \For{each epoch}
        \State {\tiny$\bullet$} Sampling via Algorithm~\ref{alg:p-sampling}.
        \For{each update step}
            \State $(s, a, c, s^\prime, u, c^p, x^\prime) \sim \mathcal{D}$

            \State {\tiny$\bullet$} Compute the state values
            
            $\qquad a_1^\prime \sim \bar{\pi}(s^\prime)$, $a_2^\prime \sim \bar{\pi}(x^\prime)$  and $u^\prime \sim \bar{\phi}(x^\prime) $
            
            $\qquad \bar{V}^\pi(x^\prime) =  \bar{Q}^\pi(x^\prime, a_2^\prime)$
            
            $\qquad \bar{V}^\pi(s^\prime) = \bar{Q}^\pi(s^\prime, a_1^\prime)$
            
            $\qquad \bar{V}^\phi(x^\prime) =  \bar{Q}^\phi(x^\prime, u^\prime)$

            \State {\tiny$\bullet$} Update the Q-networks
            
            $\qquad \psi \leftarrow \psi-\lambda_{Q} \nabla_{\psi} L_{Q^\phi} \left(\psi\right)$
            
            $\qquad \omega \leftarrow \omega-\lambda_{Q} \nabla_{\omega} L_{Q^\pi} \left(\omega\right)$

            \State {\tiny$\bullet$} Update the policy networks
            
            $\qquad \sigma \leftarrow \sigma-\lambda_{p} \nabla_{\sigma} L_{\phi} \left(\sigma\right)$
            
            $\qquad \theta \leftarrow \theta-\lambda_{p} \nabla_{\theta} L_{\pi} \left(\theta\right)$

            \State {\tiny$\bullet$} Update the target networks
            
            $\qquad \bar{\psi} \leftarrow \tau \psi+(1-\tau) \bar{\psi}$
            
            $\qquad \bar{\omega} \leftarrow \tau \omega+(1-\tau) \bar{\omega}$
            
            $\qquad \bar{\theta} \leftarrow \tau \theta+(1-\tau) \bar{\theta}$
            
            $\qquad \bar{\sigma} \leftarrow \tau \sigma+(1-\tau) \bar{\sigma}$
        \EndFor
    \EndFor
\end{algorithmic}
\end{algorithm}

\subsubsection{PR-DDPG}
We construct the four DNNs such as $Q^{\pi}(s,a;\omega)$ and $\pi(s;\theta)$ for the robust agent and $Q^{\phi}(x,u;\psi)$ and $\phi(x;\sigma)$ for the pessimistic agent. Compared with PR-DQN, the two policy networks are additionally developed for a continuous action space. During training, the network parameters $\omega$, $\theta$, $\psi$, and $\sigma$ are learned simultaneously. In PR-DDPG, the loss functions for the Q-networks are exactly same as PR-DQN, where the loss functions in \eqref{eq:pr-dqn-p}  and \eqref{eq:pr-dqn-r} are used for $Q^\phi$ and $Q^\pi$, respectively. The loss functions of the policy network for the pessimistic agent, denoted by $\phi(x;\sigma)$, defined as follows:
\begin{align} \label{eq:pr-ddpg-p}
    L_\phi(\sigma)&= \EE_{x \sim \Dc}\left[Q^\phi(x, u; \psi)\mid_{u = \phi(x ; \sigma)}\right].
\end{align} 
Also, the loss function of the policy network for the robust agent, denoted by $\pi(s;\theta)$, is defined as follows:
\begin{align} \label{eq:pr-ddpg-r}
    L_\pi(\theta)&= \EE_{s \sim \Dc}\left[Q^\pi(s, a; \omega)\mid_{a = \pi(s ; \theta)}\right].
\end{align} These loss functions are the same as the standard DDPG \cite{lillicrap2015continuous}. Nevertheless, since Q-networks are trained following the robust Bellman operator in \eqref{eq:bell3}, the robust agent in PR-DDPG can learn a robust action compared with the standard DDPG. Leveraging the updates in \eqref{eq:pr-dqn-p}, \eqref{eq:pr-dqn-r}, \eqref{eq:pr-ddpg-p}, and \eqref{eq:pr-ddpg-r}, the overall procedures of PR-DDPG are described in Algorithm \ref{alg:pr_ddpg}.

\vspace{0.2cm}
\begin{remark} While we in this paper adopted Q-Learning, DQN, and DDPG as the underlying RL algorithms, one can immediately incorporate the idea of a pessimistic agent into the other RL algorithms such as SAC \cite{haarnoja2018soft} TD3 \cite{fujimoto2018addressing} and PPO \cite{schulman2017proximal}. To be specific, the value function of a robust agent is trained using the modified loss function according to the robust Bellman operator in \eqref{eq:bell3}. Also, a pessimistic agent can be trained in the same way as PR-DQN for discrete action spaces or PR-DDPG for continuous action spaces.
\end{remark}
\vspace{0.2cm}

%
\section{Experiments}\label{sec:exp}

In this section, we provide various experimental results to validate the superiority of our algorithms. 

\subsection{Experimental Setup}

For comparison, we consider the benchmark methods below:
\begin{itemize}
\item Popular RL algorithms such as Q-Learning \cite{sutton2018reinforcement}, DQN \cite{mnih2015human}, and DDPG \cite{lillicrap2015continuous} as they are used as the underlying methods of the proposed algorithms.
\item Robust Q-Learning (in short, Robust-Q) \cite{wang2021online} as the state-of-the-art MF-RRL algorithm for tabular case.
\item To the best of our knowledge, there is no DNN-based algorithm which is constructed using the robust Bellman operator in \eqref{eq:r-contamination-bell}. In \cite{wang2021online, wang2022policy}, robust Q-Learning algorithm is developed based on a simple linear function approximation. Thus, we implemented Robust-DQN and Robust-DDPG (in short, R-DQN and R-DDPG) by appropriately combining DQN and DDPG with the robust Bellman operator in \eqref{eq:r-contamination-bell}. 
\end{itemize} 
In comparison with R-DQN and R-DDPG, we can identify the advantage of our uncertainty set compared with the state-of-the-art one in \cite{wang2021online}. Note that in these algorithms, to solve the maximization in \eqref{eq:r-contamination-bell} using samples, we stored action values in a separate buffer during training. Then, the maximum value was approximately obtained by taking the highest value from this buffer.

\begin{table}[t]
    \centering
     \caption{Hyper-parameters for Tabular Methods}
    \label{table1}
        \begin{tabular}{ c c c c c} 
         \toprule
         \bfseries  & \bfseries FrozenLake-v1 & \bfseries CliffWalking-v0\\
         \midrule 
         {$\gamma$} & 0.99 & 0.99 \\
         $\alpha$ & 0.01 & 0.01 \\
         {Batch Size} & 32 & 32 \\
         {Buffer Size} & 20000 & 20000 \\
         {Max Episode times} & 4000 & 1000 \\
         \bottomrule
        \end{tabular}
\end{table}

In our experiments, agents are trained in a training environment and then they are evaluated in a testing environment. To assess the robustness, the testing environment is constructed by intentionally adding some perturbations to the training environment. Accordingly, they have some mismatches. Among various types of perturbations (or mismatches), we take into account action and parameter perturbations.  In the action perturbations, an agent can take a random action with some probability, instead of solely adhering to an optimal policy provided by a learned agent. This type of perturbation can occur in various real-world applications. For example, when an agent's observation differs from the environment's original state, the action taken by the agent can be different from an optimal one. Also, it is possible that the agent's action can be affected by some noise in real-world environments. Regarding the parameter perturbations, some model-parameter settings in the testing environment can be different from those in the training environment. This type of perturbations can well-represent the modeling mismatches between the simulator (i.e., the training environment) and the real-world environment (i.e., the testing environment).

For experiments, we considered various environments in OpenAI Gym \cite{brockman2016openai}. Agents are trained with five instances for each algorithm and environment. At evaluation time step, we take the average reward and the standard deviation of 100 tests for each instance. In each figure, the thick line represents the average value and the shaded region shows the $\pm 0.5$ standard deviation. For tabular methods, the hyperparameter parameters provided in Table I. Also, for DNN-based MF-RRL methods, we used the hyperparameters provided by RL Baselines3 Zoo \cite{rl-zoo3}.

\subsection{Comparison with Benchmark Methods}

\begin{figure}[!t]
    \centering
    \subfigure[FrozenLake-v1(8$\times$8)]{
        \includegraphics[width=0.47\linewidth]{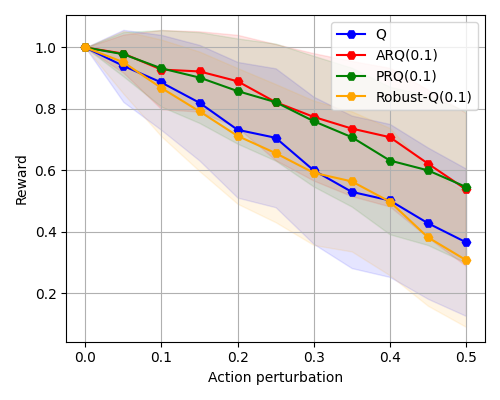} 
    }\hfill
    \subfigure[Cliffwalking-v0]{
        \includegraphics[width=0.47\linewidth]{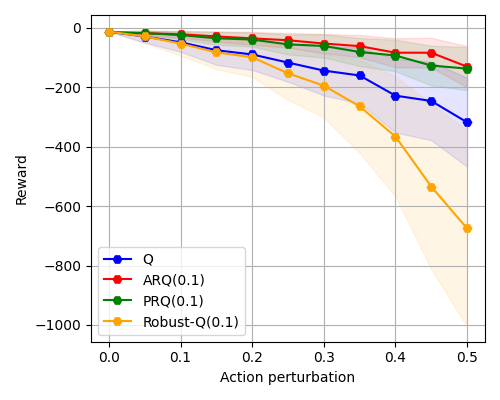}
    }
    \caption{Performance comparisons of various tabular methods under action perturbations, where FrozenLake-v1 is set as non-slippery. The values in the parenthesis indicate the robustness level $R$.} \label{fig:action_tabular}
\end{figure}

\begin{figure}[!t]
    \centering
    \subfigure[Q-Learning]{
        \includegraphics[width=0.47\linewidth]{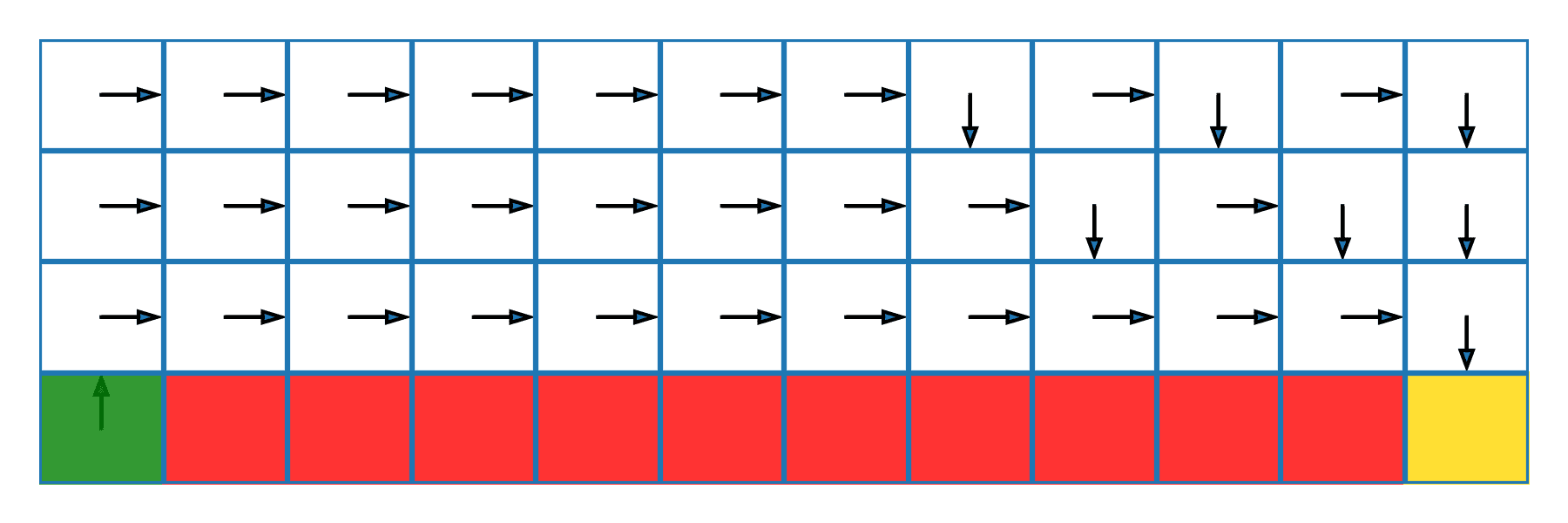} 
    }\hfill
    \subfigure[ARQ-Learning]{
        \includegraphics[width=0.47\linewidth]{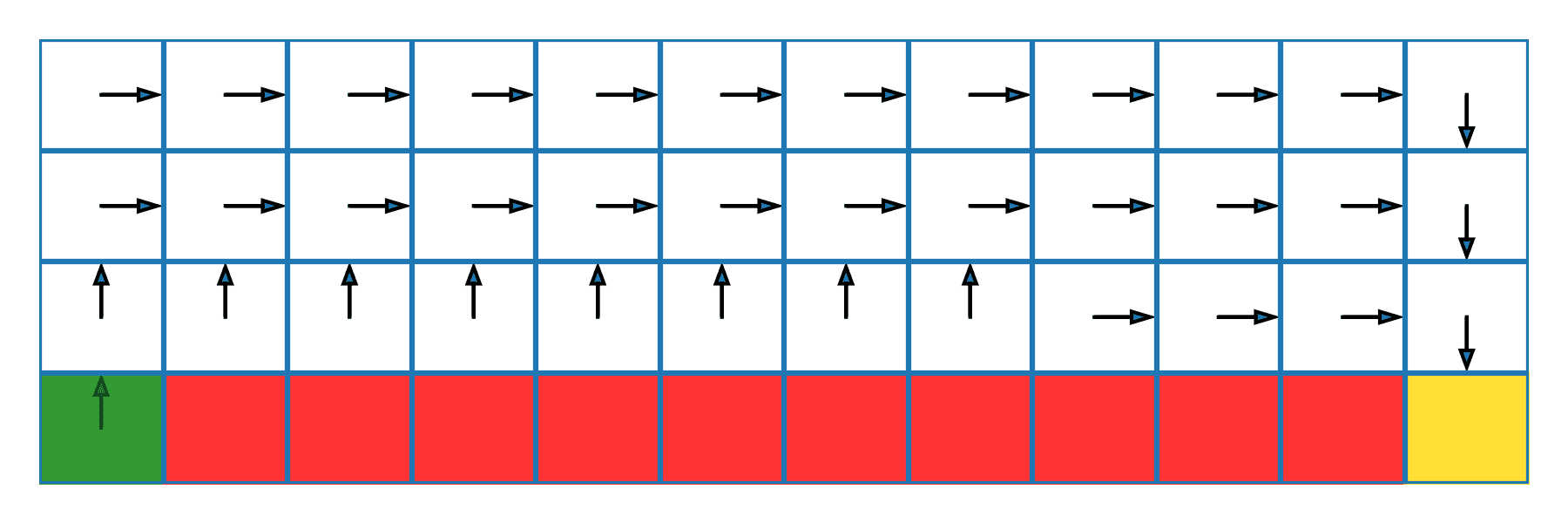}
    }
    \subfigure[PRQ-Learning]{
        \includegraphics[width=0.47\linewidth]{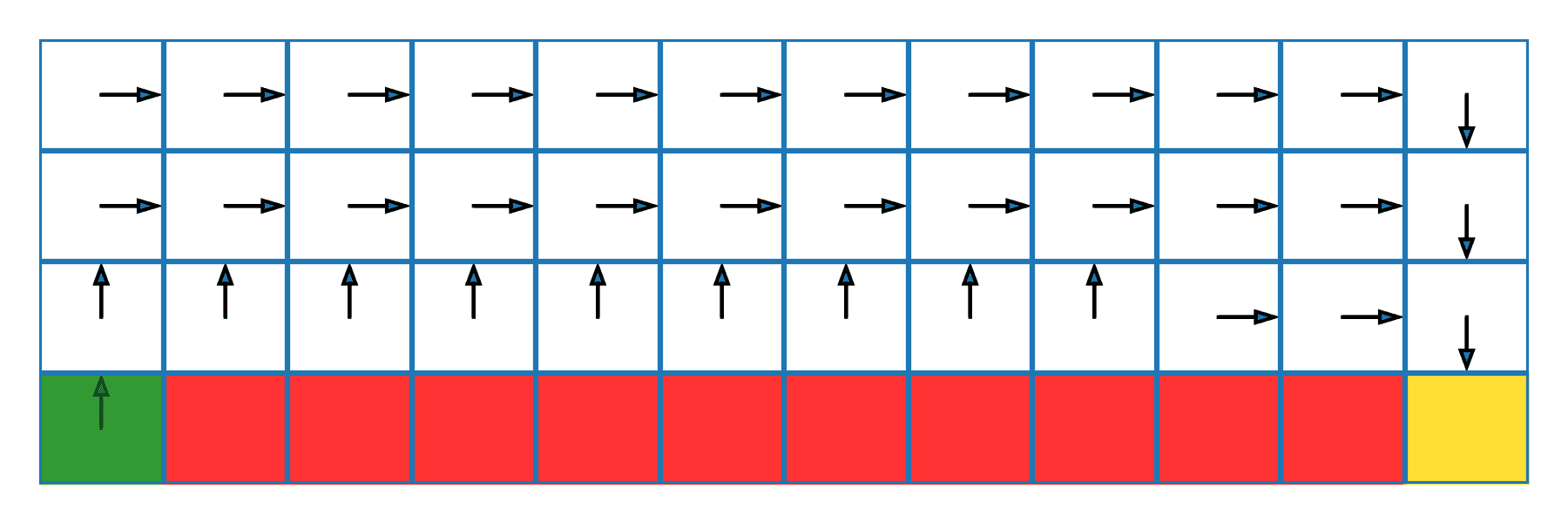}
    }\hfill
    \subfigure[robust-Q-Learning]{
        \includegraphics[width=0.47\linewidth]{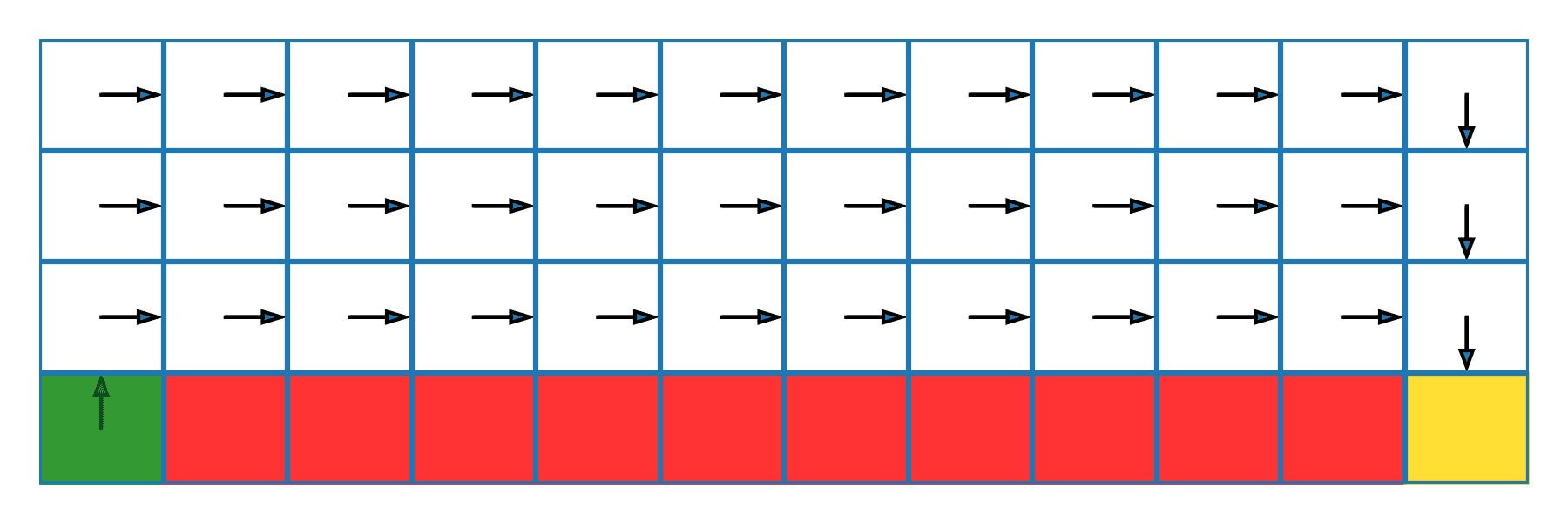}
    }
    \caption{The expressions of the optimal actions in CliffWalking-v0, where green, red, and yellow colors denote start, cliff, and goal states, respectively.}\label{fig:action_cliff}
\end{figure}

Fig.~\ref{fig:action_tabular} shows that the proposed ARQ-Learning and PRQ-Learning ensure the robustness against action perturbations, whereas the benchmark methods seem to be failed. We can identify the reason from Fig.~\ref{fig:action_cliff}, in which the red boxes represent the cliff (i.e., the dangerous states) that imposes the highest cost. It is observed that the agent trained by Q-Learning chose the actions so that it moves right above the cliff, i.e., the state transitions associated with the shortest path to reach the goal are selected. On the other hand, the agents trained by ARQ-Learning and PRQ-Learning selected the actions to result in the state transitions associated with the {\em safety} path from the cliff. These results demonstrate that as we intended, the agents in our algorithms can learn some potential dangers of adjacent (or neighboring) states. However, the agent trained with Robust-Q in \cite{wang2021online} is failed to learn a robust action for the following reason. The $R$-contamination uncertainty set for Robust-Q includes some implausible state transitions, from which the agent overestimated the potential dangers in the environment instead of reflecting them at the states nearby the cliff. Thus, our uncertainty set is more sophisticated than the state-of-the-art $Q$-contamination uncertainty set in \cite{wang2021online} in a practical way.

Focusing on continuous sate and/or action spaces, we conducted the experiments for DNN-based algorithms. The corresponding results are provided in Figs.~\ref{fig:action_perturb}$-$\ref{fig:mujoco_train_reward}. It is observed that for some cases, the proposed PR-DQN and PR-DDPG can achieve higher rewards than the benchmark methods in the presence of action and parameter perturbations. The standard RL algorithms can perform better than our algorithms in non-perturbation situations. This is natural consequence because the training and the testing environments are identical. As in the tabular case, the proposed algorithms can outperform the benchmark methods (e.g., R-DQN and R-DDPG) by ensuring better robustness to the perturbations in the testing environments. These experimental results strengthen our argument that the proposed uncertainty set in \eqref{eq:proposed_uncertainty} is elaborated than the existing one in \eqref{eq:r-contamination} in a more practical way.

\begin{figure}[!t]
    \centering
    \subfigure[ARQ-Learning]{
        \includegraphics[width=0.47\linewidth]{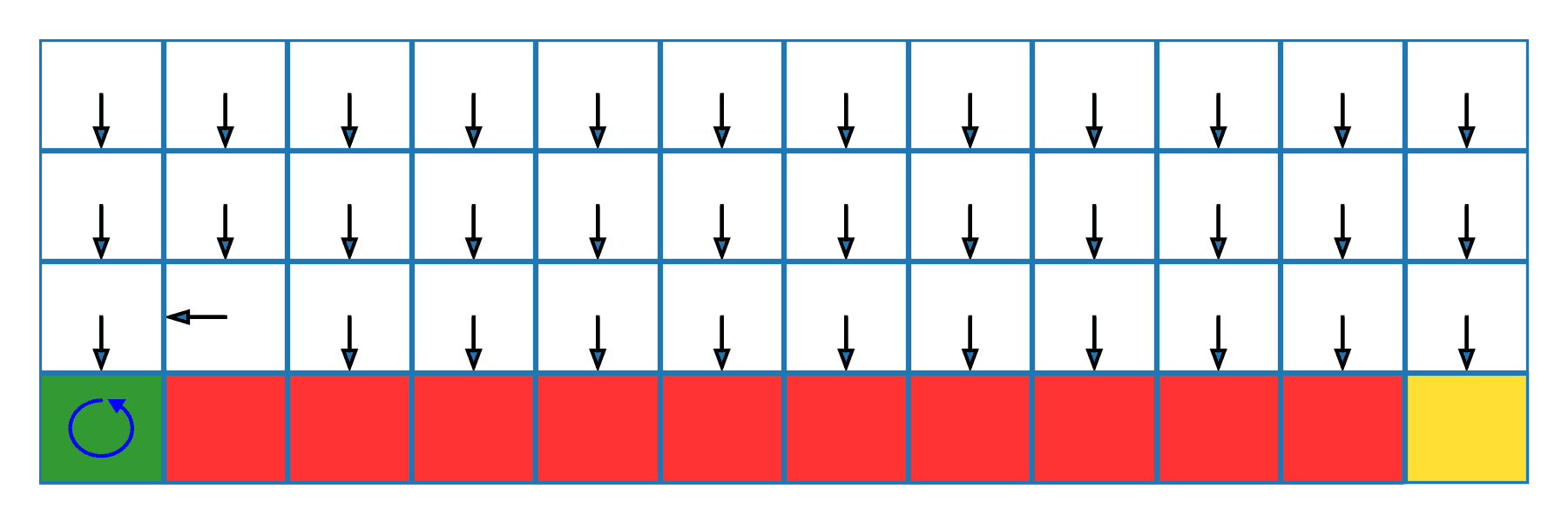} 
    }\hfill
    \subfigure[PRQ-Learning]{
        \includegraphics[width=0.47\linewidth]{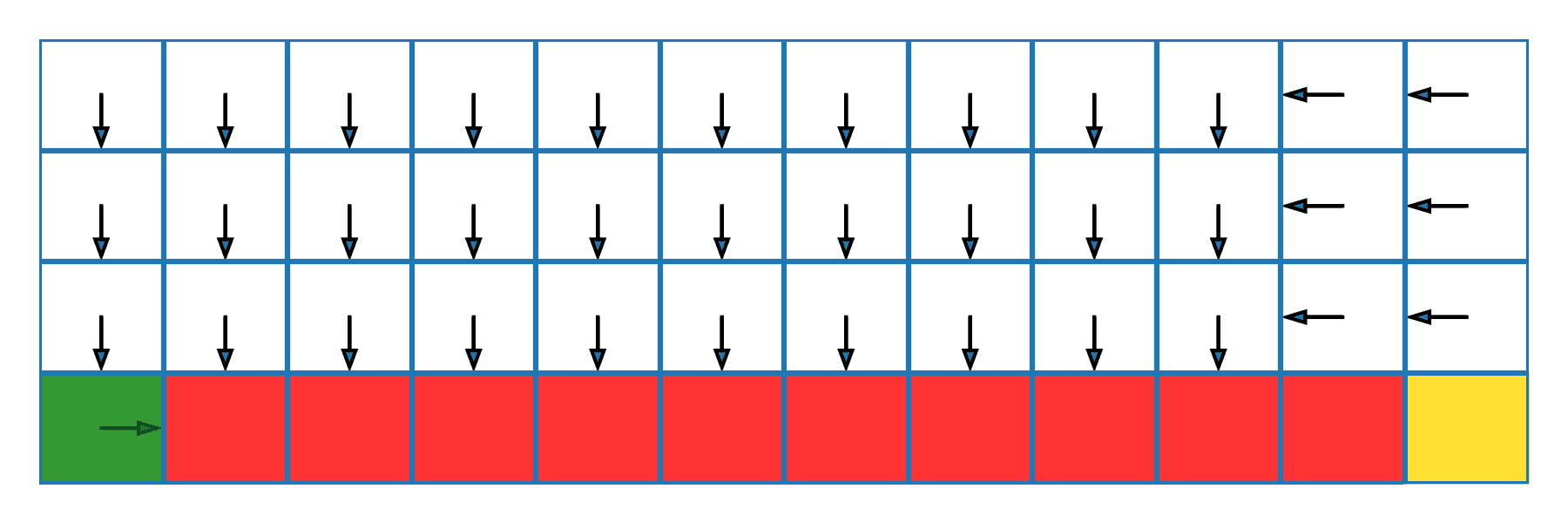}
    }
    \caption{The expressions of max-state $\hat{s}_{\rm max} = \argmax _{s'\in \Nc_s } V^\pi(s^\prime)$ provided by ARQ-Learning and PRQ-Learning in CliffWalking-v0.}
    \label{fig:pess_action_cliff}
\end{figure}

\begin{figure*}[!t]
    \centering
    \subfigure[Acrobot-v1]{
        \includegraphics[width=0.31\linewidth]{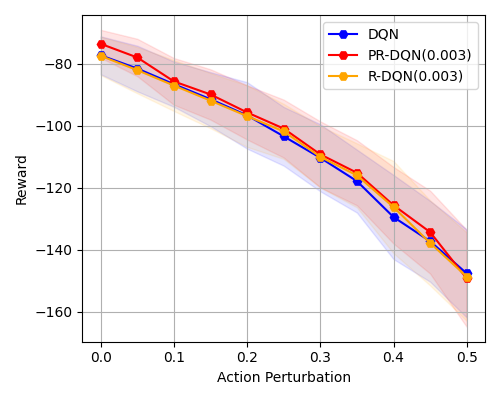} 
    }\hfill
    \subfigure[CartPole-v1]{
        \includegraphics[width=0.31\linewidth]{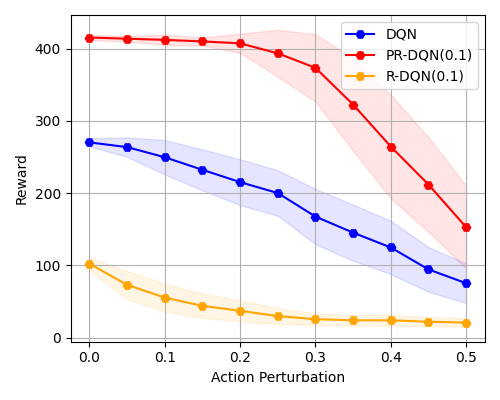}
    }\hfill
    \subfigure[MountainCar-v0]{
        \includegraphics[width=0.31\linewidth]{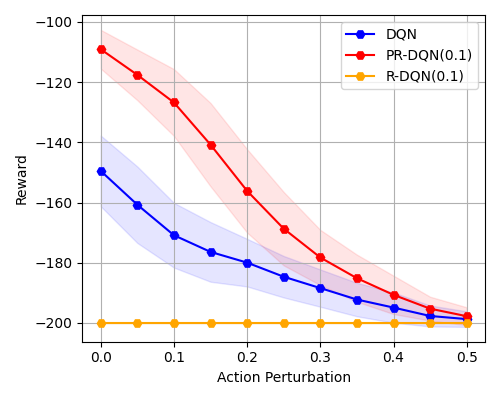}
    }
    \caption{Performance comparisons of DQN based algorithms under action perturbations in classic control environments. The values in the parenthesis indicate the robustness level $R$.}\label{fig:action_perturb}
\end{figure*}

\begin{figure*}[!t]
    \centering
    \subfigure[Acrobot-v1]{
        \includegraphics[width=0.31\linewidth]{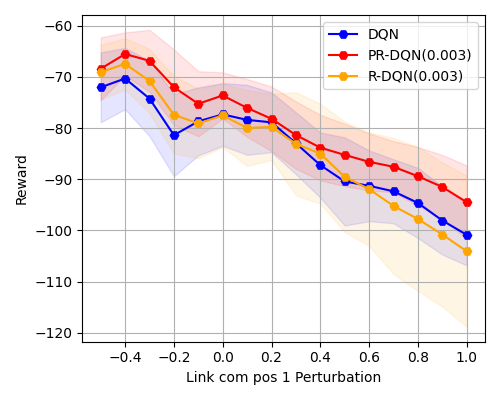} 
    }\hfill
    \subfigure[CartPole-v1]{
        \includegraphics[width=0.31\linewidth]{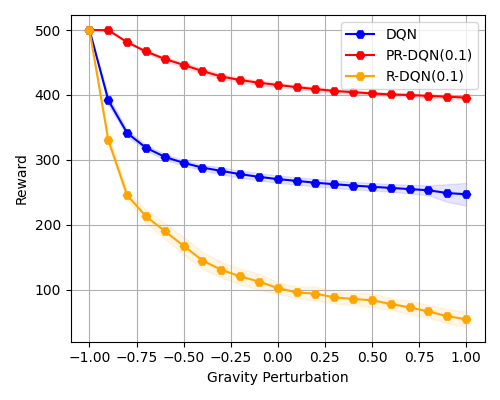}
    }\hfill
    \subfigure[MountainCar-v0]{
        \includegraphics[width=0.31\linewidth]{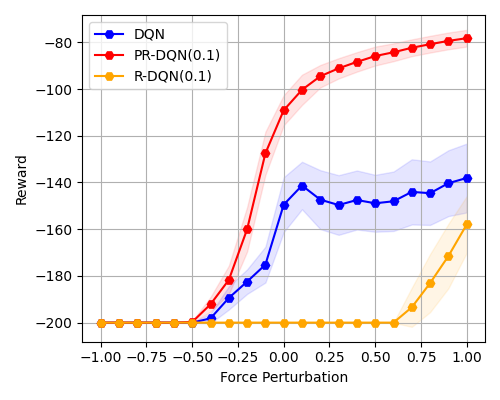}
    }
    \caption{Performance comparisons of DQN based algorithms under parameter perturbations in classic control environments. The values in the parenthesis indicate the robustness level $R$.}
    \label{fig:parameter_perturb1}
\end{figure*}

\begin{figure*}[!t]
    \centering
    \subfigure[Walker2d-v4]{
        \includegraphics[width=0.31\linewidth]{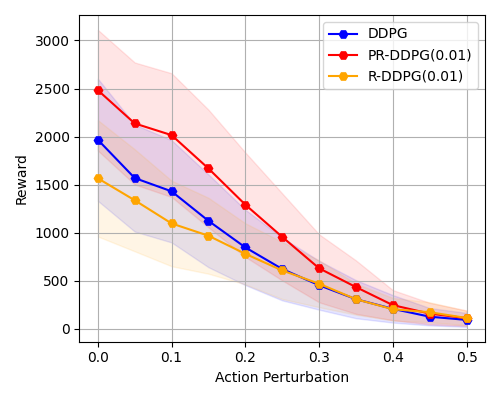} 
    }\hfill
    \subfigure[Hopper-v4]{
        \includegraphics[width=0.31\linewidth]{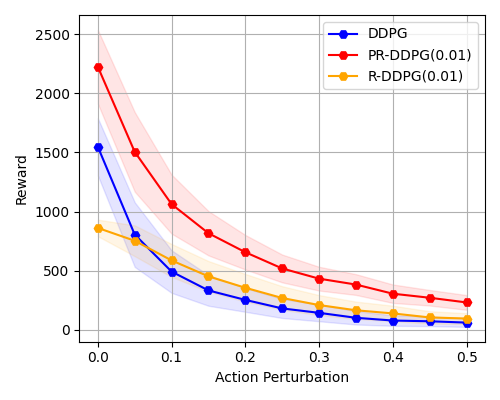}
    }\hfill
    \subfigure[HalfCheetah-v4]{
        \includegraphics[width=0.31\linewidth]{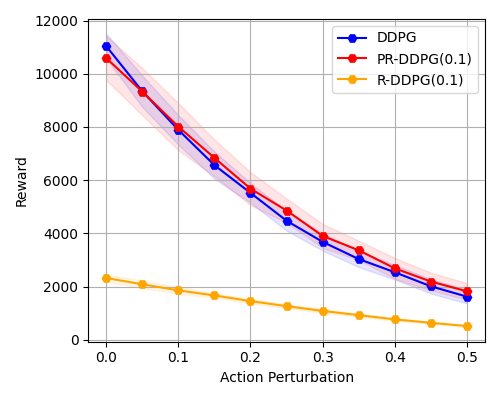}
    }
    
    \caption{Performance comparisons of DDPG based algorithms under action perturbations in MuJoCo environments \cite{todorov2012mujoco}. The values in the parenthesis indicate the robustness level $R$.}\label{fig:mujoco_train_reward}
\end{figure*}

\begin{figure*}[!t]
    \centering
    
    \subfigure[Acrobot-v1]{
        \includegraphics[width=0.31\linewidth]{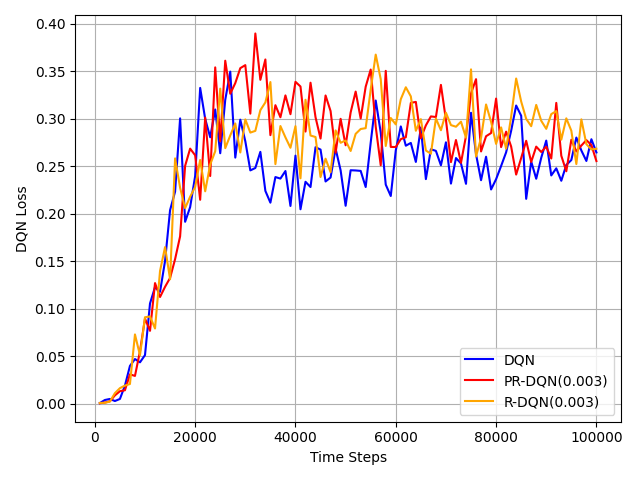} 
    }\hfill
    \subfigure[CartPole-v1]{
        \includegraphics[width=0.31\linewidth]{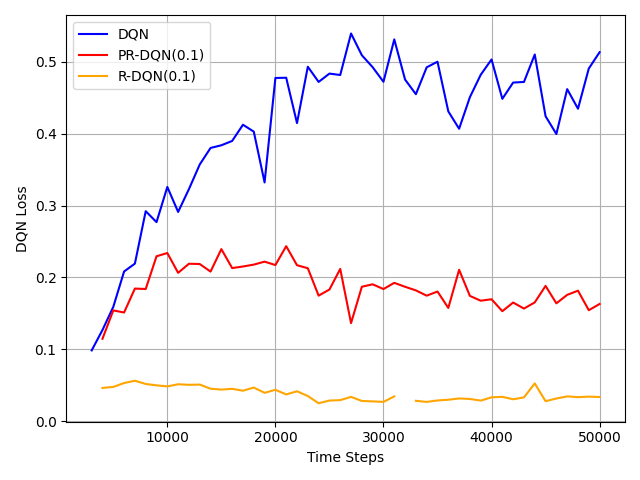}
    }\hfill
    \subfigure[MountainCar-v0]{
        \includegraphics[width=0.31\linewidth]{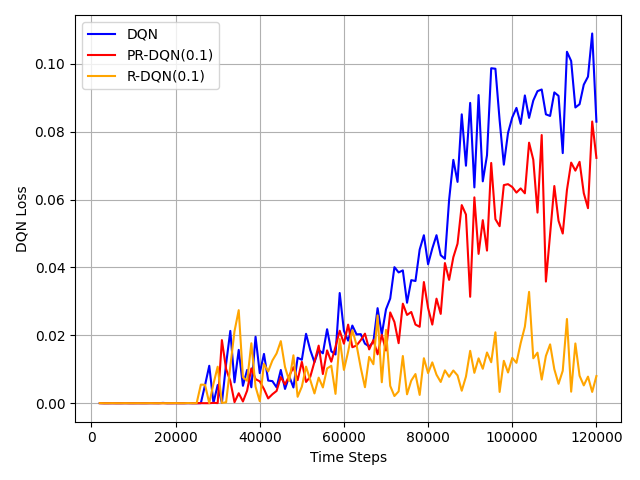}
    }

    \caption{The loss curves for DQN during training in classic control environments. The values in the parenthesis indicate the robustness level $R$.}\label{fig:class_control_train_loss}
\end{figure*}

\begin{figure*}[!t]
    \centering
    
    \subfigure[Walker2d-v4]{
        \includegraphics[width=0.31\linewidth]{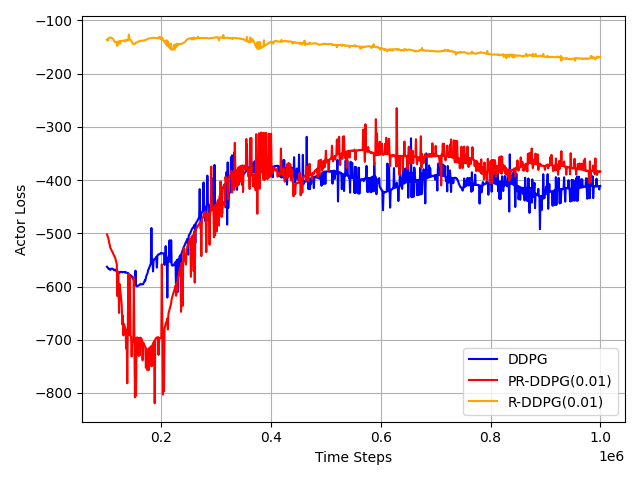} 
    }\hfill
    \subfigure[Hopper-v4]{
        \includegraphics[width=0.31\linewidth]{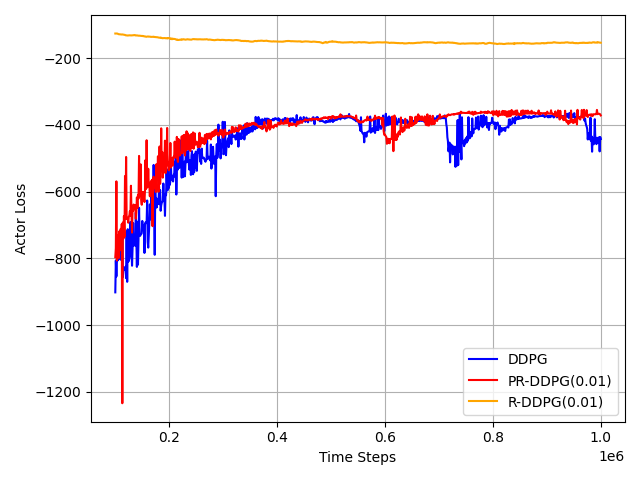}
    }\hfill
    \subfigure[HalfCheetah-v4]{
        \includegraphics[width=0.31\linewidth]{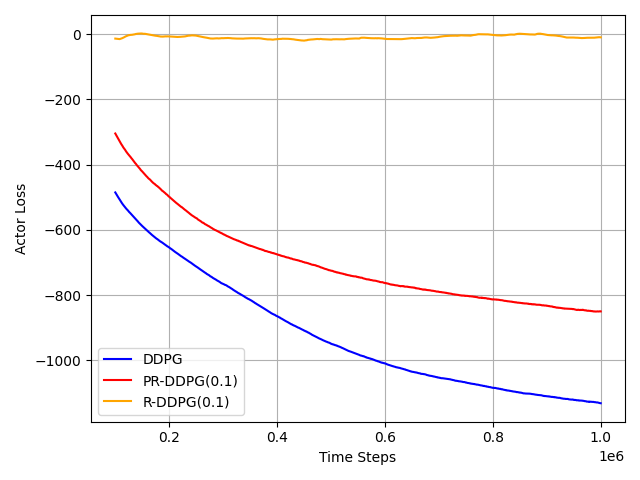}
    }
    
    \subfigure[Walker2d-v4]{
        \includegraphics[width=0.31\linewidth]{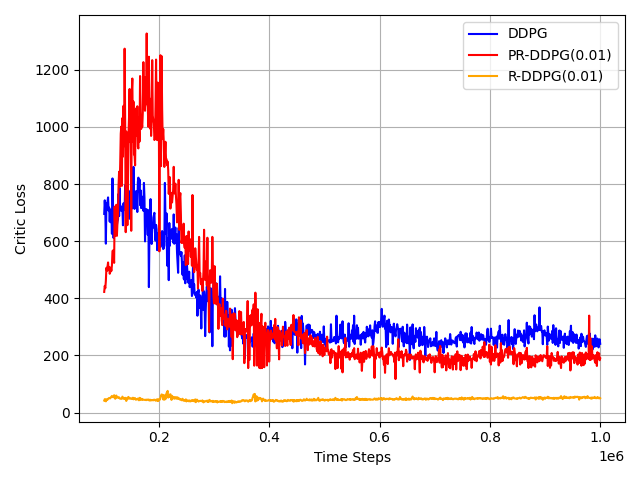} 
    }\hfill
    \subfigure[Hopper-v4]{
        \includegraphics[width=0.31\linewidth]{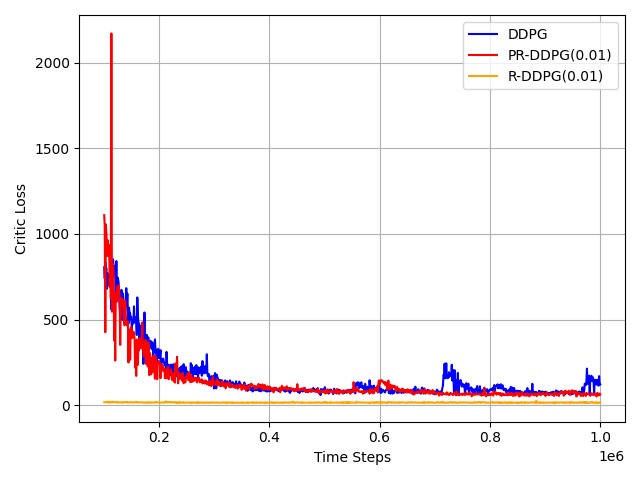}
    }\hfill
    \subfigure[HalfCheetah-v4]{
        \includegraphics[width=0.31\linewidth]{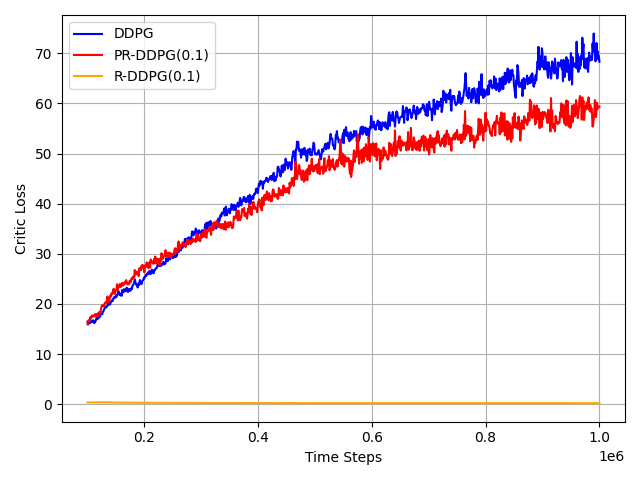}
    }
    \caption{The loss curves for actor and critic networks in MuJoCo environments \cite{todorov2012mujoco}. The values in the parenthesis indicate the robustness level $R$.}\label{fig:mujoco_train_loss}
\end{figure*}

%
\subsection{Ablation Study}\label{ablation}

Via experiments, we verify the effectiveness of our major contributions such as adjacent $R$-contamination uncertainty set and a pessimistic agent.

\noindent{{\bf The accuracy of pessimistic agents.}} In Fig.~\ref{fig:pess_action_cliff}, an estimated max-state $\hat{s}_{\rm max} \in \Sc=\{\mbox{left}, \mbox{right}, \mbox{up}, \mbox{down}\}$ is depicted, where
\begin{equation*}
    \hat{s}_{\rm max}=\argmax_{s' \in \Nc_{s}} V^{\pi}(s').
\end{equation*} It is observed that the estimated max-states in ARQ-Learning and PRQ-Learning are almost identical. This confirmed that our pessimistic agent indeed solves the maximization in in the robust Bellman operator in \eqref{eq:bell2}. 

\noindent{{\bf The risk of $R$-contamination uncertainty set in \cite{wang2021online}.}} From Figs.~\ref{fig:action_perturb}$-$\ref{fig:mujoco_train_reward}, it becomes apparent that R-DQN and R-DDPG based on the $R$-contamination uncertainty set had difficult in learning a robust policy. In addition, Figs.~\ref{fig:class_control_train_loss} and~\ref{fig:mujoco_train_loss} show that the loss of R-DQN and R-DDPG is significantly lower than the proposed PR-DQN and PR-DDPG, respectively. Namely, both R-DQN and R-DDPG are not learned well. As identified in Section \ref{subsec:challenges}, MF-RRL is highly vulnerable to outliers as the worst-case performance is optimized. One can speculate that the presence of outliers in the $R$-contamination uncertainty set reduced the reward values, consequently hindering the learning. Due to this issue, furthermore, both R-DQN and R-DDPG can lack stability and thus, their performances would be significantly affected by variations in the uncertainty set. In contrast, the proposed PR-DQN and PR-DDPG can ensure stability as our uncertainty set is constructed by excluding such outliers from the $R$-contamination uncertainty set.

\begin{figure*}[!t]
    \centering
    \subfigure[Walker2d-v4]{
        \includegraphics[width=0.31\linewidth]{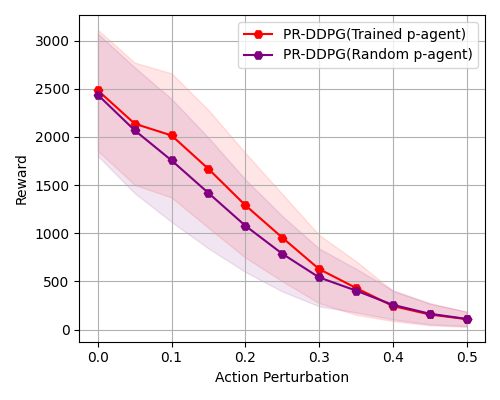} 
    }\hfill
    \subfigure[Hopper-v4]{
        \includegraphics[width=0.31\linewidth]{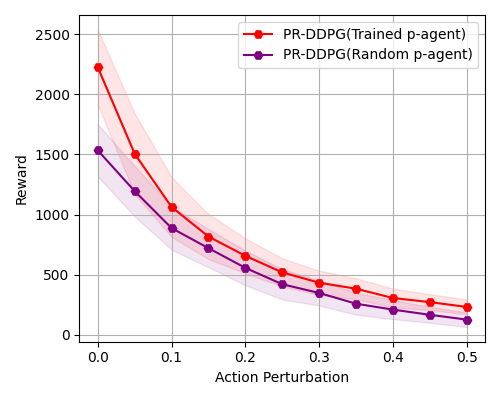}
    }\hfill
    \subfigure[HalfCheetah-v4]{
        \includegraphics[width=0.31\linewidth]{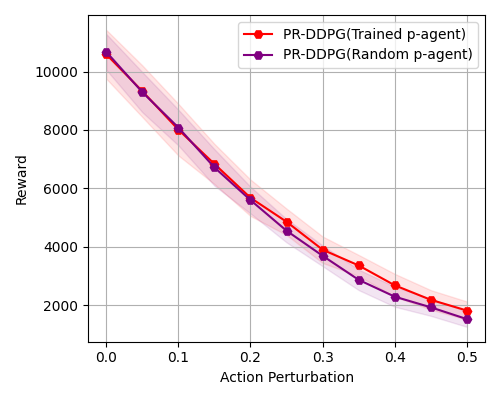}
    }

    \caption{Performance comparisons between the proposed pessimistic agent and the random pessimistic agent. }\label{fig:mujoco_ablation}
\end{figure*}

\begin{figure*}[!t]
    \centering
    
    \subfigure[Walker2d-v4]{
        \includegraphics[width=0.31\linewidth]{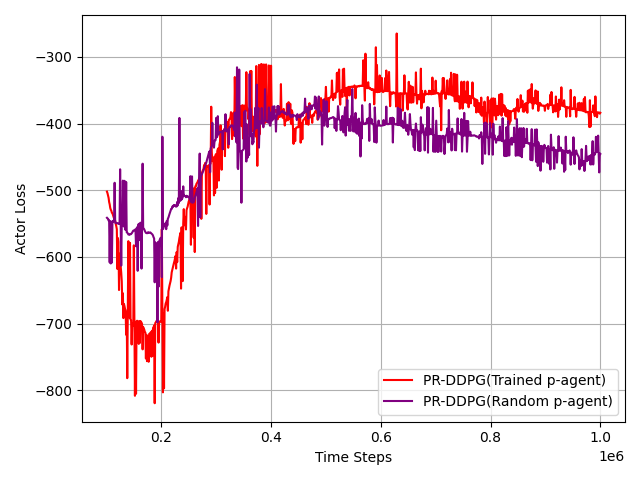} 
    }\hfill
    \subfigure[Hopper-v4]{
        \includegraphics[width=0.31\linewidth]{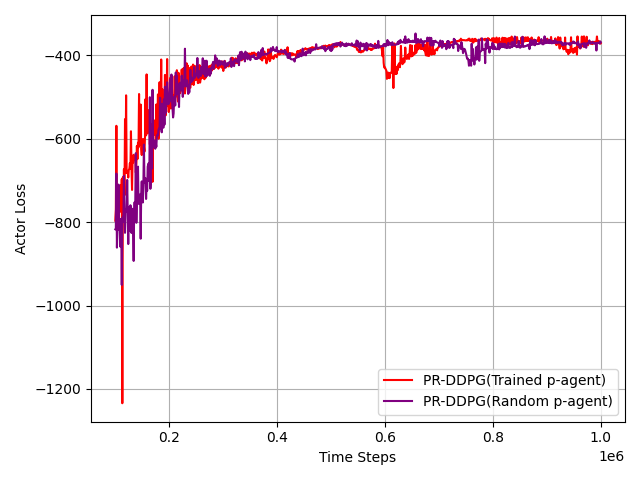}
    }\hfill
    \subfigure[HalfCheetah-v4]{
        \includegraphics[width=0.31\linewidth]{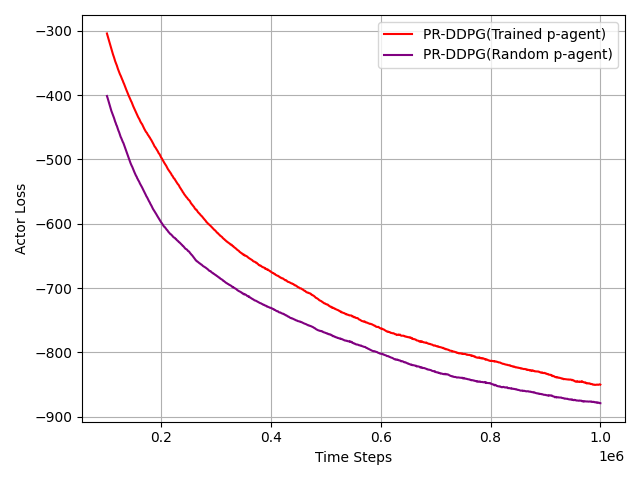}
    }
    
    \subfigure[Walker2d-v4]{
        \includegraphics[width=0.31\linewidth]{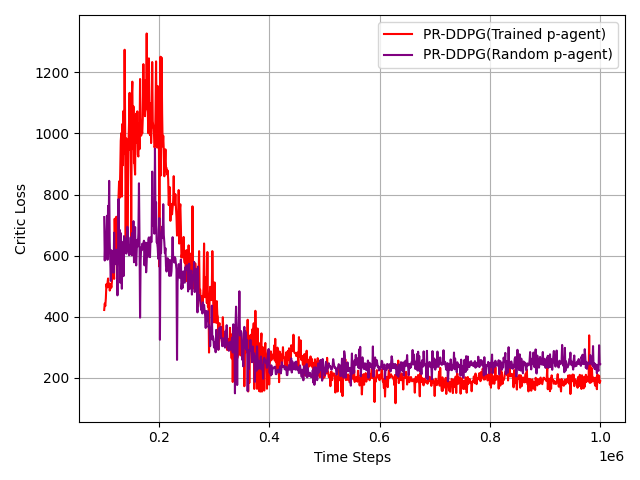} 
    }\hfill
    \subfigure[Hopper-v4]{
        \includegraphics[width=0.31\linewidth]{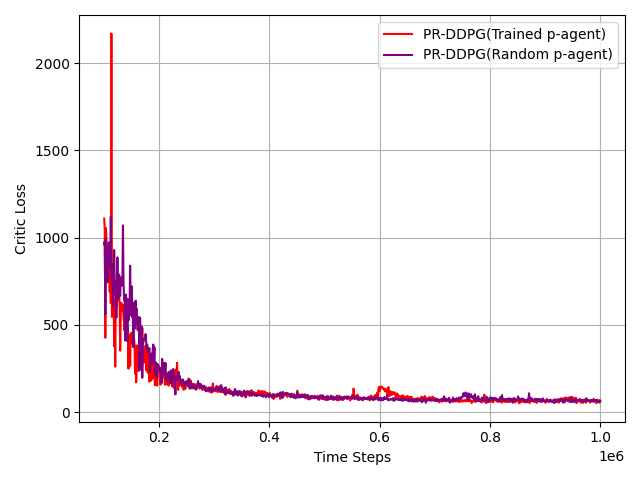}
    }\hfill
    \subfigure[HalfCheetah-v4]{
        \includegraphics[width=0.31\linewidth]{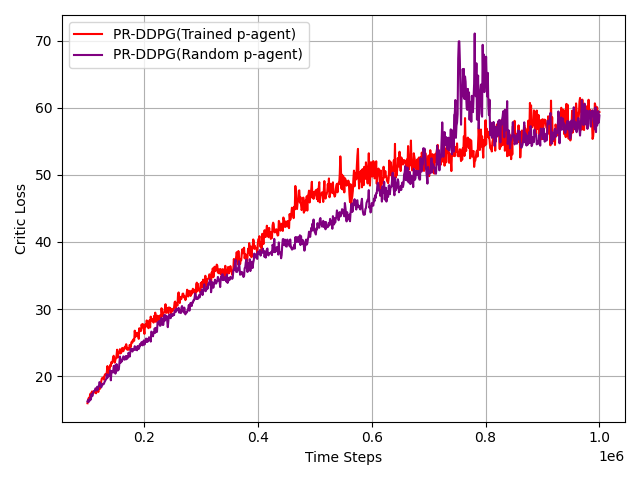}
    }
    \caption{The loss comparisons between the proposed pessimistic agent and the random pessimistic agent.}\label{fig:mujoco_ablation_loss}
\end{figure*}

\noindent{{\bf The stability of pessimistic agents.}} We conducted the experiments to verify the stability of our pessimistic agent in terms of yielding a good solution to the maximization in \eqref{eq:pess_condition1}. One might concern that the pessimistic agent cannot find an optimal solution especially in the beginning of training, namely,
\begin{equation}\label{eq:not}
   x' \neq \argmax_{s'\in\Nc_s} V^{\pi}(s'),
\end{equation} where $x'$ is the estimated solution by the pessimistic agent. We emphasize that our pessimistic agent at least guarantees that 
\begin{equation}\label{eq:c1}
    x' \in \Nc_s\subseteq \Sc.
\end{equation} In contrast, the condition in \eqref{eq:c1} is not guaranteed when the $R$-contamination uncertainty set in \cite{wang2021online} is adopted. We argue that due to such difference, the proposed algorithms can provide better performances than the benchmark methods. In the experiments to verify this, we introduced the so-called {\em random} pessimistic agent which is not trained and selects a state $x' \in \Nc_s$ uniformly and randomly as the solution of the maximization. This can mimic a poorly trained pessimistic agent. From Figs. \ref{fig:mujoco_ablation} and \ref{fig:mujoco_ablation_loss}, it is demonstrated that PR-DDPG with the random pessimistic agent can attain a similar performance with the standard PR-DDPG. This is because the proposed uncertainty set is designed to be more stable, excluding implausible state transitions (i.e., outliers).
Although it is happened at some time steps that the pessimistic agent dose not find an exact solution of the maximization in \eqref{eq:pess_condition1}, the proposed PR-DDPG can be well-trained, thereby yielding an attractive performance.

\begin{table}[t]
    \centering
     \caption{Comparisons of Computational Complexity\\ (Total Runtme$(s)$)}
    \label{table:cost}
        \begin{tabular}{ c c c c c} 
         \toprule
          & \bfseries DDPG & \bfseries PR-DDPG & \bfseries R-DDPG \\
         \midrule 
         Walker2d-v4 & 11414.8 & 28619.4 & 14324.8  \\
         Hopper-v4 & 16914.8 & 26845 & 13823.4 \\
         HalfCheetah-v4 & 14512.4 & 27027.8 & 13693.2 \\
         \bottomrule
        \end{tabular}
\end{table}

\subsection{Computational Costs}\label{cost_comparision}

We compare the computational costs of our algorithms with the benchmark methods in MuJoCo environments \cite{todorov2012mujoco}. As a metric, we evaluate the total runtime to learn an algorithm (i.e., the required time for training). The corresponding results are provided in Table~\ref{table:cost}. As expected, the proposed DDPG takes approximately twice the time compared to the standard DDPG and R-DDPG. This is because our algorithm needs to train the two agents (i.e., the robust and the pessimistic agents) and each agent requires sampling from the training environment, whereas the other algorithms need to train the robust agent only. If our runtime becomes a serious problem, it can be further reduced via parallel processing. Namely, learning the robust and the pessimistic agents can be conducted in parallel.

\section{Conclusion}\label{sec:con}

We presented a new uncertainty set, named adjacent $R$-contamination uncertainty set, which better reflects real-world perturbations than the existing uncertainty sets. Based on this set, we developed ARQ-Learning for tabular cases and characterized its performance bound theoretically. It was also proved that ARQ-Learning can converge to an optimal value as fast as Q-Learning and the state-of-the-art robust Q-Learning while ensuring better robustness. We introduced an additional pessimistic agent, which can tackle the major bottleneck for the extension of ARQ-Learning into large or continuous state spaces. Leveraging the idea of the pessimistic agent, we developed PRQ-Learning suitable for tabular cases with large state spaces, and PR-DQN and PR-DDPG for continuous state spaces. While the proposed algorithms are based on Q-Learning, DQN, and DDPG as the underlying RL algorithms, we emphasize that our idea can be immediately integrated with the other RL algorithms such as SAC, TD3, and PPO. Via experiments, we verified the effectiveness of our algorithms on various RL applications with model mismatches. It would be an interesting future work to extend our idea into offline RL.

\appendices
\section{Proof of Theorem 1}\label{proof1}
We prove Theorem 1. In \cite{wang2021online}, it was proved that the robust Bellman operator for $R$-contamination uncertainty set is contraction. In this section, we extend this to the proposed {\em adjacent} $R$-contamination uncertainty set. In the below, it is assumed that the neighboring set of a state $s \in \Sc$ at least includes $s$ as its element $s \in \Nc_s$, to ensure that the neighboring set is not empty set (i.e., $\Nc_s \ne  \emptyset$). This assumption makes sense in practice because if an agent does not take any action, it can stay at the current state with non-zero probability. First we have:
\begin{align}
    &|\Tm Q_1(s, a) - \Tm Q_2(s, a)| \nonumber\\
    = & \Bigg|c(s, a) + \gamma(1-R)\left[\sum_{s^\prime \in S} \bar{p}_{s, s^\prime}^a V_1(s^\prime) \right] + \gamma R \left[\max _{s'\in  \Nc_s } V_1(s^\prime)\right] \nonumber \\
     - & c(s, a) - \gamma(1-R)\left[\sum_{s^\prime \in S} \bar{p}_{s, s^\prime}^a V_2(s^\prime) \right] - \gamma R \left[\max _{s'\in  \Nc_s } V_2(s^\prime)\right] \Bigg| \nonumber\\
    = & \bigg| \gamma(1 - R)\left[\sum_{s^\prime \in S} \bar{p}_{s, s^\prime}^a (V_1(s^\prime) - V_2(s^\prime)) \right] \nonumber \\
     &\quad +\gamma R\left(\max _{s'\in  \Nc_s } V_1(s^\prime) - \max _{s'\in  \Nc_s } V_2(s^\prime)\right)\bigg| \nonumber\\
    \le & \left|\gamma(1 - R)\left[\sum_{s^\prime \in S} \bar{p}_{s, s^\prime}^a \left(\min_a Q_1(s^\prime, a) - \min_a Q_2(s^\prime, a) \right) \right] \right| \nonumber \\
    & \quad +\left| \gamma R\max _{s'\in  \Nc_s } \left(V_1(s^\prime) - V_2(s^\prime)\right)\right| \nonumber\\
    \le & \gamma(1 - R) \sum_{s^\prime \in S} \bar{p}_{s, s^\prime}^a \left|\min_a Q_1(s^\prime, a) - \min_a Q_2(s^\prime, a) \right| \nonumber \\
    &\quad +\gamma R\max _{s'\in  \Nc_s } \left|V_1(s^\prime) - V_2(s^\prime)\right| \nonumber\\
    \stackrel{(a)}{\leq} & \gamma(1 - R)\left\|Q_1 - Q_2\right\|_\infty  + \gamma R \left\|Q_1 - Q_2\right\|_\infty \nonumber\\
    = & \gamma\left\|Q_1 - Q_2\right\|_\infty,
\end{align} where (a) is due to the following facts 1) and 2): 
\begin{enumerate}
\item If $\min_a Q_1(s^\prime, a) \ge \min_a Q_2(s^\prime, a)$, then we have:
\begin{align*}
    &\left|\min_a Q_1(s^\prime, a) - \min_a Q_2(s^\prime, a) \right| \\
    & \quad\quad = \min_a Q_1(s^\prime, a) - \min_a Q_2(s^\prime, a) \nonumber\\ 
    & \quad\quad  \le Q_1(s^\prime, b) -  Q_2(s^\prime, b) \nonumber\\
    & \quad\quad \le \left\|Q_1-Q_2\right\|_{\infty},
\end{align*} where $b = \argmin_a Q_2(s^\prime, a)$.
\item If $\min_a Q_1(s^\prime, a) < \min_a Q_2(s^\prime, a)$, then we have:
\begin{align*}
      &\left|\min_a Q_1(s^\prime, a) - \min_a Q_2(s^\prime, a) \right| \\
      & \quad\quad = \min_a Q_2(s^\prime, a) - \min_a Q_1(s^\prime, a) \nonumber\\ 
      & \quad\quad \le Q_2(s^\prime, c) -  Q_1(s^\prime, c) \nonumber\\
     &\quad\quad \le \left\|Q_1-Q_2\right\|_{\infty},
\end{align*} where $c = \argmin_a Q_1(s^\prime, a)$. 
\end{enumerate}
Also, we can obtain the following inequality:
\begin{align}
    \max _{s'\in  \Nc_s } \left|V_1(s^\prime) - V_2(s^\prime)\right| \le & \max _{s'\in \mathcal{S} } \left|V_1(s^\prime) - V_2(s^\prime)\right| \nonumber \\
    \le & \left\|Q_1 - Q_2\right\|_\infty.
\end{align} We now proved that our robust Bellman operator $\Tm$ is contraction with respect to $l_\infty$-norm. From Banach fixed point theorem \cite{puterman2014markov}, we also proved that $Q^\pi$ has the unique fixed point of $\Tm$. This completes the proof of Theorem 1.

\section{Proof of Theorem 2}\label{proof2}

We prove Theorem 2. As in Appendix A, it is assumed that an estimated neighboring set  $\hat{\Nc}_s$ is not empty set (i.e., $\hat{\Nc}_s \ne \emptyset$). As explained before, it is reasonable assumption in practical RL applications. We first introduce the useful vector notations. Let $V_t \in 
\RR^{|\mathcal{S}|}$ and $Q_t \in \RR^{|\mathcal{S}||\mathcal{A}|}$ denote the vector representations of the estimation of value functions at time step $t$. Also, we let $V^* \in \RR^{|\mathcal{S}|}$ and $Q^* \in \RR^{|\mathcal{S}||\mathcal{A}|}$ denote the vector representations of optimal value functions. Let  $c \in \RR^{|\mathcal{S}||\mathcal{A}|}$ be the vector representation of the cost function such that the $(s,a)$-th entry of $c$ is equal to $c(s,a)$. 

Using the vector notations, given $(s,a)\in \Sc\times\Ac$, we define the transition kernel vector $P^{a}_{s} \in \RR^{|\mathcal{S}|}$ as
\begin{equation}
    P_s^a(s^\prime) = \bar{p}_{s, s^\prime}^a.
\end{equation}
Also, given a sample $O_t = (s_t, a_t, s_{t+1})$, we can define the sample transition vector $P_{s, t+1}^a \in \RR^{|\mathcal{S}|}$ as
\begin{equation}
    P_{s, t+1}^a(s^\prime ) = \begin{cases} 1, & \mbox{if } (s, a, s^\prime) = O_t \\
    0, & \mbox{othewise.}
\end{cases}
\end{equation}
Then, the update of ARQ-Learning at the $(s, a)$-th element can be rewritten as
\begin{align}
     Q_t(s, a)  &= (1 - \alpha \mathbbm{1}_{\{s = s_t\}})Q_{t-1}(s, a) \nonumber \\
    & \quad + \alpha\mathbbm{1}_{\{s = s_t\}}\Big(c(s,a) + \gamma(1 - R)(P_{s, t}^{a})^{\trasp}V_{t-1}  \nonumber\\
    &\quad\quad\quad\quad\quad\quad\quad\quad\quad + \gamma R\max_{s' \in \hat{\Nc}_{s_t}}V_{t - 1}(s') \Big)\label{eq:q_bellman_1} \\
    & = (1 - \alpha \mathbbm{1}_{\{s = s_t\}})Q_{t-1}(s, a) \nonumber \\ 
    & \quad + \alpha \mathbbm{1}_{\{s = s_t\}}\Big(c(s,a) + \gamma(1 - R)(P_{s, t}^{a})^{\trasp}V_{t-1} \nonumber \\
    &\quad\quad\quad\quad\quad\quad\quad\quad\quad + \gamma R\max_{s' \in \hat{\Nc}_{s}}V_{t - 1}(s') \Big), \label{eq:q_bellman_2}
\end{align}
where $\mathbbm{1}_{\{\cdot\}}$ represents an indicator function. Although $\hat{\Nc}_{s_t}$ and $\hat{\Nc}_{s}$ are different in the maximization in \eqref{eq:q_bellman_1} and \eqref{eq:q_bellman_2}, we can easily identify that \eqref{eq:q_bellman_2} is equivalent to \eqref{eq:q_bellman_1} due to the indicator function $\mathbbm{1}_{\{s = s_t\}}$. Also, the optimal robust Bellman equation can be expressed as
\begin{equation}
    Q^{\star}(s, a) = c(s, a) + \gamma(1 - R)(P_s^a)^{\trasp}V^{\star} + \gamma R\max_{s'\in {\Nc}_{s}}V^{\star}(s').
\end{equation} We let:
\begin{equation}
\psi_t(s, a) \eqdef Q_t(s, a) - Q^{\star}(s, a).
\end{equation}
Using this notation, we have:
\begin{align*}
    \psi_t(s, a) &= Q_t(s, a) - Q^{\star}(s, a) \nonumber \\
    &= (1 - \alpha \mathbbm{1}_{\{s = s_t\}})\left(Q_{t-1}(s,a) - Q^{\star}(s,a)\right) \nonumber \\
    &\;\; + \alpha \mathbbm{1}_{\{s = s_t\}}(Q_t(s,a) - Q^{\star}(s,a)) \nonumber \\ 
    &= (1 - \alpha \mathbbm{1}_{\{s = s_t\}})\psi_{t-1}(s, a) \nonumber \\
    & \;\; + \alpha \mathbbm{1}_{\{s = s_t\}}\gamma(1 - R)\left((P_{s, t}^{a})^{\trasp}V_{t-1} - (P_s^a)^\trasp V^{\star}\right) \nonumber \\ 
    & \;\; + \alpha \mathbbm{1}_{\{s = s_t\}} \gamma R\left(\max_{s' \in \hat{\Nc}_{s}}V_{t - 1}(s') - \max_{s'\in \Nc_{s}}V^{\star}(s')\right) \nonumber \\
    &= (1 - \alpha \mathbbm{1}_{\{s = s_t\}})\psi_{t-1}(s, a) \nonumber \\
    &\;\; + \alpha\gamma(1 - R) \mathbbm{1}_{\{s = s_t\}}\left((P_{s, t}^{a})^\trasp V_{t-1} - (P_{s, t}^{a})^\trasp V^{\star} \right) \nonumber \\
    & \;\; + \alpha\gamma(1 - R) \mathbbm{1}_{\{s = s_t\}}\left( (P_{s,t}^{a})^\trasp V^{\star} - (P_{s}^{a})^\trasp V^{\star}\right) \nonumber\\ 
    & \;\; + \alpha\gamma R \mathbbm{1}_{\{s = s_t\}}\left(\max_{s' \in \hat{\Nc}_{s}}V_{t - 1}(s') - \max_{s' \in \Nc_{s}}V^{\star}(s') \right) \nonumber \\
    &= (1 - \alpha \mathbbm{1}_{\{s = s_t\}})\psi_{t-1}(s,a) \nonumber \\
    & \;\; + \alpha\gamma(1 - R) \mathbbm{1}_{\{s = s_t\}}\left(P_{s,t}^{a} - P_{s}^{a}\right)^\trasp V^{\star} \nonumber\\
    &\;\; + \alpha\gamma(1 - R) \mathbbm{1}_{\{s = s_t\}}(P_{s, t}^{a})^\trasp (V_{t-1} - V^{\star}) \nonumber \\
    & \;\; + \alpha\gamma R \mathbbm{1}_{\{s = s_t\}}\left(\max_{s' \in \hat{\Nc}_{s}}V_{t - 1}(s') - \max_{s' \in \Nc_{s}}V^{\star}(s')\right).
\end{align*}
Applying this relation recursively, we can get:
\begin{equation}
    \psi_t(s,a) = \beta_{1, t}(s,a) + \beta_{2, t}(s,a) + \beta_{3, t}(s,a),\label{eq:beta}
\end{equation}
where $\beta_{1, t}(s,a), \beta_{2, t}(s,a)$, and $\beta_{3, t}(s,a)$ are respectively defined as 
\begin{align}
    \beta_{1, t}(s,a)&=\prod_{j=1}^t\left(1- \mathbbm{1}_{\{s = s_j\}}\right) \psi_0 (s, a)  \\
    \beta_{2, t}(s,a)&=\gamma(1-R) \sum_{i=1}^t \prod_{j=i+1}^t\left(1-\mathbbm{1}_{\{s = s_j\}}\right) \mathbbm{1}_{\{s = s_i\}} \nonumber \\
     & \qquad \qquad \quad\quad\quad  \times \left(P_{s,i}^{a}-P_{s}^{a}\right)^\trasp V^{\star}   \\
    \beta_{3, t}(s,a)&=\gamma(1-R) \sum_{i=1}^t \prod_{j=i+1}^t\left(1-\mathbbm{1}_{\{s = s_j\}}\right) \mathbbm{1}_{\{s = s_i\}} \nonumber \\
    & \qquad \qquad \quad\quad\quad \times (P_{s,i}^{a})^\trasp \left(V_{i-1}-V^{\star}\right) \nonumber \\
    & \qquad +\gamma R \sum_{i=1}^t \prod_{j=i+1}^t\left(1-\mathbbm{1}_{\{s = s_j\}}\right) \mathbbm{1}_{\{s = s_i\}} \nonumber \\ 
    & \qquad \qquad \times \left(\max _{s \in \hat{\Nc}_{s}} V_{i-1}(s)-\max _{s \in \hat{\Nc}_{s}} V^{\star}(s)\right).
\end{align}
Applying the triangle inequality to \eqref{eq:beta}, we can obtain the upper bound of $\psi_t(s,a)$ such as
\begin{equation}
    |\psi_t(s,a)| \le |\beta_{1, t}(s, a)| + |\beta_{2, t}(s, a)| + |\beta_{3, t}(s, a)|.
\end{equation} In Lemmas 1, 2, and 3 below, we derive the upper bounds of $\beta_{1, t}(s,a), \beta_{2, t}(s,a)$, and $\beta_{3, t}(s,a)$, respectively. Combining them, we can obtain the upper bound of $\psi_t(s, a)$ as follows:
\begin{itemize}
\item For $t<t_{\text {frame }}$, we have:
\begin{align*}
    \left|\psi_t(s,a)\right| \leq & \left\|\psi_0\right\|_{\infty} + \gamma c \sqrt{\alpha \log \frac{|\mathcal{S}||\mathcal{A}|T}{\delta}}\left\|V^{\star}\right\|_{\infty} \nonumber \\ 
     & + \gamma \sum_{i=1}^t\left\|\psi_{i-1}\right\|_{\infty} \prod_{j=i+1}^t\left(1-\mathbbm{1}_{\{s = s_j\}}\right) \mathbbm{1}_{\{s = s_i\}}.
\end{align*}
\item For $t_{\text {frame}} \leq t \leq T$, we have:
\begin{align*}
    \left|\psi_t(s,a)\right| &\leq  (1-\alpha)^{\frac{t \mu_{\min }}{2}}\left\|\psi_0\right\|_{\infty} \nonumber \\
    & + \gamma c \sqrt{\alpha \log \frac{|\mathcal{S} \| \mathcal{A}|T}{\delta}}\left\|V^{\star}\right\|_{\infty} \nonumber \\ 
    & +\gamma \sum_{i=1}^t\left\|\psi_{i-1}\right\|_{\infty} \prod_{j=i+1}^t\left(1-\mathbbm{1}_{\{s = s_j\}}\right) \mathbbm{1}_{\{s = s_i\}}.
\end{align*}
\end{itemize}
This bound exactly matches the bound in the Equation (42) in \cite{li2020sample}. Thus, the remaining part of Theorem 2 can be obtained by exactly following the proof of \cite[Theorem 5]{li2020sample}. This completes the proof of Theorem 2.

\vspace{0.5cm}
\begin{lemma} For any $\delta>0$ and $t_{\text {frame }} =\frac{443 t_{\text {mix }}}{\mu_{\text {min }}} \log \left(\frac{4|\mathcal{S}||\mathcal{A}| T}{\delta}\right)$ with $ t_{\text {frame }} \le T$,  with probability at least $1-\delta$, $\beta_{1, t}(s,a)$ is bounded as
\begin{equation*}
    \left|\beta_{1, t}(s,a)\right| \leq
    \begin{cases}
    (1-\alpha)^{\frac{1}{2} t \mu_{\min }}\left\|\psi_0 \right\|_{\infty},\; t_{\text {frame }} < \forall t < T\\
    \left|\beta_{1, t}(s,a)\right| \leq \left\|\psi_0 \right\|_{\infty},\; \forall t < t_{\text {frame }}.
    \end{cases} 
\end{equation*}
\end{lemma}
\begin{IEEEproof}
    The proof exactly follows the proof of \cite[Lemma 2]{li2020sample}.
\end{IEEEproof}

\vspace{0.3cm}
\begin{lemma}  There exists some constant $c>0$ such that for any $0<\delta<1$ and $t \le T$ that satisfies $0<\alpha \log \frac{|\mathcal{S} \| \mathcal{A}| T}{\delta}<1$, with probability at least $1-\delta$, we have that for $\forall (s, a) \in \mathcal{S} \times \mathcal{A}$, 
\begin{align*}
    \left|\beta_{2, t}(s,a)\right| & \leq c \gamma (1-R) \sqrt{\eta \log \left(\frac{|\mathcal{S}||\mathcal{A}| T}{\delta}\right)}\left\|V^{\star}\right\|_{\infty} \nonumber \\
    & \le c \gamma \sqrt{\eta \log \left(\frac{|\mathcal{S}||\mathcal{A}| T}{\delta}\right)}\left\|V^{\star}\right\|_{\infty}.
\end{align*}
\end{lemma}
\begin{IEEEproof}
    The proof exactly follows the proof of \cite[Lemma 1]{li2020sample}.
\end{IEEEproof}

\vspace{0.3cm}
\begin{lemma} For any $t > 0$, we have that for $\forall (s, a) \in \mathcal{S} \times \mathcal{A}$, 
\begin{equation*}
    |\beta_{3, t}(s, a)| \le \gamma \sum_{i = 1}^{t} \left\|\psi_{t-1} \right\|_\infty \prod_{j = i+1}^t \left(1-\mathbbm{1}_{\{s = s_j\}}\right) \mathbbm{1}_{\{s = s_i\}}.
\end{equation*}
\end{lemma}
\begin{IEEEproof}
We first have:
\begin{align}
    \left|(P_{s, i}^{a})^\trasp (V_{i-1} - V^{\star})\right| & \le \left\|P_{s, i}^{a}\right\|_\infty\left\|V_{i-1} - V^{\star}\right\|_\infty \nonumber \\
    & \le \left\|V_{i-1}-V^{\star}\right\|_{\infty} \nonumber \\ 
    & =\max _s\left|V_{i-1}(s)-V^{\star}(s)\right| \nonumber \\
    &=\left|V_{i-1}\left(s^{\star}\right)-V^*\left(s^{\star}\right)\right| \nonumber \\
    & =\left|\min _a Q_{i-1}\left(s^{\star}, a\right)-\min _b Q^{\star}\left(s^{\star}, b\right)\right| \nonumber \\ 
    & \leq\left\|Q_{i-1}-Q^{\star}\right\|_{\infty} \nonumber \\ 
    &= \left\|\psi_{i-1}\right\|_{\infty},
\end{align} where $s^{\star}=\arg\max_{s \in \Sc} \left|V_{i-1}(s)-V^{\star}(s)\right|$. Also, we can get:
\begin{align}
  & \left|\max _{s' \in \hat{\Nc}_{s}} V_{i-1}(s')-\max _{s' \in \Nc_{s}} V^{\star}(s') \right| \\ 
  \stackrel{(a)}{\le} &\left| \max_{s^\prime\in \hat{\mathcal{N}}_s} V_{i-1}(s^\prime) - \max_{s^\prime \in \hat{\mathcal{N}}_s}V^{\star}(s^\prime) \right| \\
    & +  \left| \max_{s^\prime
\in \hat{\mathcal{N}}_s} V^{\star}(s^\prime) - \max_{s^\prime
\in \mathcal{N}_s} V^{\star}(s^\prime) \right| \\ 
 \stackrel{(b)}{=} & \left| \max_{s^\prime\in\hat{\mathcal{N}}_s} V_{i-1}(s^\prime) - \max_{s^\prime \in \hat{\mathcal{N}}_s}V^{\star}(s^\prime) \right| \\
\le & \left\|V_{i-1} - V^{\star} \right\|_{\infty} \nonumber \\ 
    \leq & \left\|Q_{i-1}-Q^{\star}\right\|_{\infty} \nonumber \\
    =  & \left\|\psi_{i-1}\right\|_{\infty},
\end{align} where (a) is due to the triangle inequality and (b) is from Assumption 2. It is obvious that $0 \le \left(1-\mathbbm{1}_{\{s = s^\prime\}}\right) \le 1$ and $0 \le \mathbbm{1}_{\{s = s^\prime\}} \le 1$ for $\forall s, s^\prime \in \mathcal{S}$. Thus, we have:
\begin{equation*}
    |\beta_{3, t}(s, a)| \le \gamma \sum_{i = 1}^{t} \left\|\psi_{t-1} \right\|_\infty \prod_{j = i+1}^t \left(1-\mathbbm{1}_{\{s = s_j\}}\right) \mathbbm{1}_{\{s = s_i\}}.
\end{equation*} This completes the proof of Lemma 3.

\end{IEEEproof}

\ifCLASSOPTIONcaptionsoff
  \newpage
\fi


\bibliographystyle{ieeetr}
\bibliography{main_rev}

\end{document}